\documentclass[acmtog,nonacm]{acmart}

\usepackage{multirow}
\usepackage{booktabs} 
\usepackage{url}
\usepackage{amsmath,bm}
\usepackage[ruled]{algorithm2e} 

\SetAlFnt{\small}
\SetAlCapFnt{\small}
\SetAlCapNameFnt{\small}
\SetAlCapHSkip{0pt}
\setcopyright{none}
\usepackage{overpic} 
\usepackage{subcaption}

\DeclareMathAlphabet{\mathcal}{OMS}{cmsy}{m}{n}

\acmJournal{TOG}

\usepackage{array}
\newcolumntype{L}[1]{>{\raggedright\let\newline\\\arraybackslash\hspace{0pt}}m{#1}}
\newcolumntype{C}[1]{>{\centering\let\newline\\\arraybackslash\hspace{0pt}}m{#1}}
\newcolumntype{R}[1]{>{\raggedleft\let\newline\\\arraybackslash\hspace{0pt}}m{#1}}

\newcommand{\Tref}[1]{Table~\ref{#1}}

\newcommand{\Fref}[1]{Figure~\ref{#1}}

\begin{document}
\title{Cross-Domain and Disentangled Face Manipulation with 3D Guidance}

\author{Can Wang}
\affiliation{
  \institution{City University of Hong Kong}
}
\email{cwang355-c@my.cityu.edu.hk}
\author{Menglei Chai}
\affiliation{
  \institution{Snap Inc.}
}
\email{cmlatsim@gmail.com}
\author{Mingming He}
\affiliation{
  \institution{USC Institute for Creative Technologies}
}
\email{hmm.lillian@gmail.com}
\author{Dongdong Chen}
\affiliation{
  \institution{Microsoft Cloud AI}
}
\email{cddlyf@gmail.com}
\author{Jing Liao$^*$}\thanks{*Corresponding Author}
\affiliation{
  \institution{City University of Hong Kong}
}
\email{jingliao@cityu.edu.hk}

\begin{abstract}
Face image manipulation via three-dimensional guidance has been widely applied in various interactive scenarios due to its semantically-meaningful understanding and user-friendly controllability. However, existing 3D-morphable-model-based manipulation methods are not directly applicable to out-of-domain faces, such as non-photorealistic paintings, cartoon portraits, or even animals, mainly due to the formidable difficulties in building the model for each specific face domain. To overcome this challenge, we propose, as far as we know, the first method to manipulate faces in arbitrary domains using human 3DMM. This is achieved through two major steps: 1) disentangled mapping from 3DMM parameters to the latent space embedding of a pre-trained StyleGAN2~\cite{DBLP:conf/cvpr/KarrasLAHLA20} that guarantees disentangled and precise controls for each semantic attribute; and 2) cross-domain adaptation that bridges domain discrepancies and makes human 3DMM applicable to out-of-domain faces by enforcing a consistent latent space embedding. Experiments and comparisons demonstrate the superiority of our high-quality semantic manipulation method on a variety of face domains with all major 3D facial attributes controllable – pose, expression, shape, albedo, and illumination. Moreover, we develop an intuitive editing interface to support user-friendly control and instant feedback. Our project page is \textcolor{black}{\url{https://cassiepython.github.io/cddfm3d/index.html}}
\end{abstract}

%
%

\ccsdesc[500]{Computing methodologies~Image processing}

\begin{teaserfigure}
\centering
\setlength{\tabcolsep}{0\linewidth}
\begin{tabular}{C{0.16\linewidth}C{0.02\linewidth}C{0.16\linewidth}C{0.16\linewidth}C{0.16\linewidth}C{0.16\linewidth}C{0.16\linewidth}}
\multicolumn{7}{c}{\includegraphics[width=0.98\textwidth]{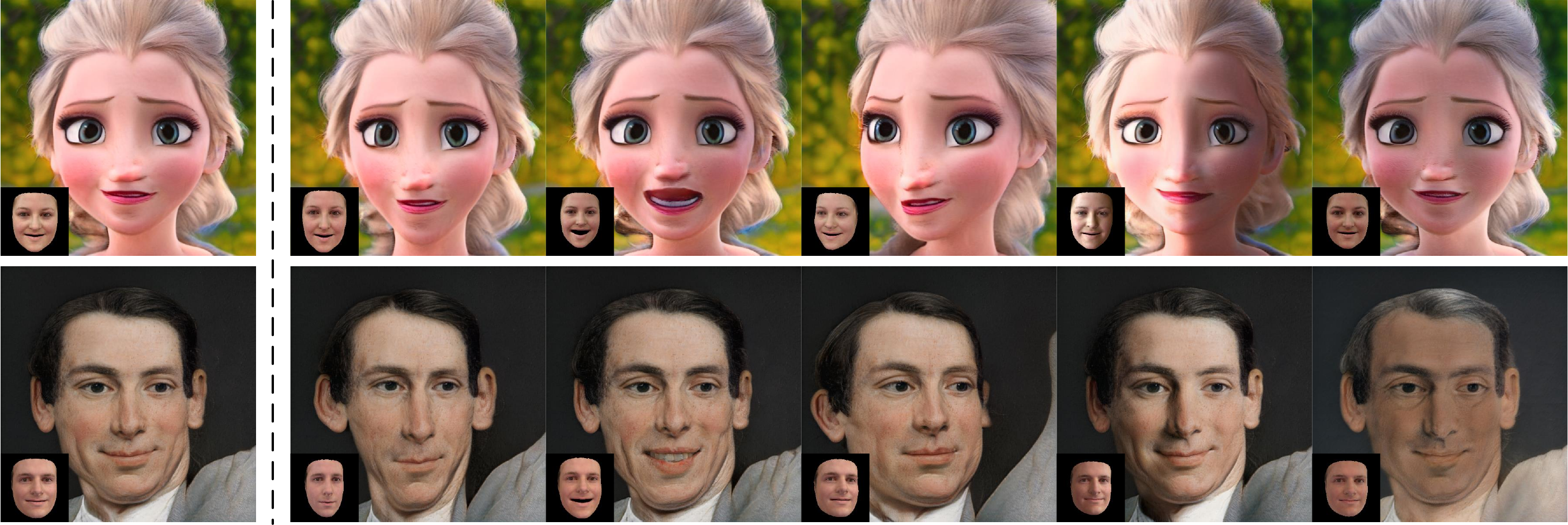}}\\
original&&shape&expression&pose&illumination&albedo\\
\end{tabular}
\caption{Our cross-domain face manipulation pipeline allows disentangled manipulation of arbitrary out-of-domain face images. Controlled by 3D human face parameters (show as the inset in each result), our method enables editing of a wide variety of semantic facial attributes, including shape, expression, pose, illumination, and albedo.}
\label{fig:teaser}
\end{teaserfigure}

\keywords{Face Image Manipulation, Domain Adaptation, 3D Morphable Model, Disentanglement, StyleGAN}

\maketitle

\section{Introduction}
Face image manipulation has been long coveted in various interactive scenarios, such as photo enhancement, social games, and virtual reality. With the striking development of human digitization techniques, 3D-guided face manipulation has popularized itself through semantically meaningful understanding and user-friendly controllability. With the strong parameterization capability of 3D morphable models (3DMMs)~\cite{DBLP:conf/siggraph/BlanzV99,DBLP:conf/cvpr/GenovaCMSVF18,DBLP:journals/tvcg/WenWRCH20,DBLP:journals/tvcg/AsthanaHDG12}, there arises the possibility of editing various facial attributes, including pose, expression, shape, albedo, and illumination. Along this direction, before the deep learning era, the problem is mainly addressed from the perspective of the traditional graphics pipeline, in which a face model is fitted to the subject in the image and then re-rendered with altered facial parameters. However, due to the limitations of the low dimensionality of 3DMM and the approximated shading models, the results often suffer from unsatisfactory realism. 
\textcolor{black}{Although some recent works \cite{DBLP:conf/cvpr/SmithSDTTE20,DBLP:conf/cvpr/LattasMGPTGZ20} improve the albedo reconstruction of 3DMM, they are still unable to produce photorealistic faces with high-frequency details due to the limited number of scans and the lack of realism of their renderings.}
More recently, the rapid advancement of deep generative networks, such as StyleGAN~\cite{DBLP:conf/cvpr/KarrasLA19} and StyleGAN2~\cite{DBLP:conf/cvpr/KarrasLAHLA20}, enables photo-realistic face synthesis. To bring the best of both worlds, the pioneering works of StyleRig~\cite{DBLP:conf/cvpr/TewariEBBSPZT20} and PIE~\cite{DBLP:journals/tog/TewariE0BSPZT20} achieve huge success via adopting 3DMM as the proxy of intuitive parametric controls and utilizing the pre-trained StyleGAN2 to synthesize the corresponding high-quality manipulation results.

Despite the progress made in 3D-guided face manipulation, most existing methods are available only for human faces and cannot be trivially extended to out-of-domain faces, such as non-photorealistic paintings, cartoon portraits, or even animals. This is mainly due to the formidable difficulties in building the 3DMM for each specific face domain regarding data acquisition and processing. To overcome this challenge, we propose, as far as we know, the first method to manipulate arbitrary out-of-domain faces via human 3DMM. This is achieved through two major steps: \textit{disentangled attribute-latent mapping} and \textit{cross-domain adaptation}.

Within the human-face domain, the disentangled attribute-latent mapping structure is adopted to bridge the latent space and the 3DMM parameters. Specifically, a source latent code, inverted from the input image, is first mapped to its 3DMM parameters for manipulation. Then, the edited parameters are projected back to update the latent code that generates the final image. Different from StyleRig~\cite{DBLP:conf/cvpr/TewariEBBSPZT20} that treats the latent code of StyleGAN2 as an indivisible element in mapping, we propose the reduced style space, which is decomposed into subspaces corresponding to different semantic 3DMM attributes (\textit{i.e.} pose, expression, shape, albedo, and illumination). Based on that, the aforementioned mapping is learned between each attribute and its corresponding subspace in a disentangled manner. Our method minimizes the mutual interference between different attributes and thus greatly improves the quality and flexibility of the manipulation.

When extending to out-of-domain faces, our cross-domain adaptation makes the semantic latent embedding consistent for all domains. Inspired by the recent observation in \cite{DBLP:arxiv/pinkney2020resolution,DBLP:arxiv/huang2020unsupervised}, we perform latent-consistent finetuning that adapts the StyleGAN2 generator to another domain while keeping the latent spaces aligned. Therefore, given an out-of-domain face image, we first optimize its corresponding latent code through cross-domain latent inversion and then manipulate it using the in-domain attribute-latent mapping that is consistent for all domains. Furthermore, when fed into the original human-face StyleGAN2 generator, the inverted latent code can be mapped to a real human face. As a manipulation proxy, any edits on it will be faithfully reflected on the out-of-domain face input by simply feeding the same edited latent code into the finetuned StyleGAN2 for out-of-domain faces, thanks to the shared latent embedding.

Our method enables high-quality semantic manipulation on a variety of out-of-domain faces with all major 3D facial attributes controllable, including pose, expression, shape, albedo, and illumination, as shown in Fig.~\ref{fig:teaser}. Empowered by StyleGAN2, these edits maintain visually realistic occlusion handling, lighting consistency, and perspective parallax, which can be hardly achieved with traditional approaches. Furthermore, with our reduced latent space and disentangled attribute-latent mapping, our method enjoys highly-disentangled manipulation, which allows it to edit one attribute without affecting irrelevant content or manipulate multiple attributes simultaneously without introducing mutual interference. Experiments and comparisons demonstrate the superiority of the proposed method compared to previous manipulation methods (either in image space \cite{DBLP:journals/tog/SchaeferMW06,DBLP:conf/nips/SiarohinLT0S19} or latent space\cite{DBLP:conf/nips/HarkonenHLP20,DBLP:conf/cvpr/ShenGTZ20}), regarding both quality and controllability. In addition, we develop an editing interface for user-friendly manipulation based on the 3D face proxy, which achieves intuitive controls and instant feedback.

The contributions of this paper can be summarized as follows:
\begin{itemize}
\item We propose the first method to manipulate semantic attributes of out-of-domain faces guided by human face 3DMM;
\item Our reduced style space and disentangled attribute-latent mapping allow disentangled and precise control over all facial attributes of 3DMM, achieving high-quality face image manipulation;
\item Our cross-domain adaptation bridges the domain discrepancy and makes human face 3DMM applicable to out-of-domain faces by enforcing a consistent latent space embedding. 
\end{itemize}

Our source code, pre-trained models, and data will be publicly available at \textcolor{black}{\url{https://cassiepython.github.io/cddfm3d/index.html}}

\section{Related Work}

\subsection{Image Space Face Manipulation}
The traditional methods for face manipulation include image deformation based on sparse control points using Moving Least Squares (MLS) ~\cite{DBLP:journals/tog/SchaeferMW06} and image warping to approximate the 3D space editing effects~\cite{DBLP:journals/tog/ShechtmanGF16,DBLP:journals/tog/Averbuch-ElorCK17}. Recently, deep neural networks have been extensively exploited
towards the warping-based synthesis of facial geometry. \textcolor{black}{Geng $\textit{et al}$} \cite{DBLP:journals/tog/GengSZWZ18} use a generative adversarial network (GAN) to synthesize appearance details onto a pre-warped face image. Wiles~\emph{et al.}~\cite{DBLP:conf/eccv/WilesKZ18} propose to generate a dense motion field by using neural networks to warp the image towards a reference face. Siarohin~\emph{et al.} ~\cite{DBLP:conf/cvpr/SiarohinLT0S19,DBLP:conf/nips/SiarohinLT0S19} propose to encode motion information via keypoints learned in a self-supervised fashion and then warp the image according to the reference keypoint trajectories. Facial landmarks are also popular to represent facial geometry, widely used in conditional face translation~\cite{DBLP:conf/accv/NatsumeYM18,DBLP:conf/iccv/NirkinKH19,DBLP:conf/iccv/ZakharovSBL19,DBLP:conf/aaai/HaKKSK20,DBLP:journals/corr/abs-2004-12452}. Conditioned on 2D landmarks, some frameworks~\cite{DBLP:conf/accv/NatsumeYM18,DBLP:conf/iccv/NirkinKH19} focus on face swapping between unseen identities, while other methods~\cite{DBLP:conf/iccv/ZakharovSBL19,DBLP:conf/aaai/HaKKSK20,DBLP:journals/corr/abs-2004-12452} achieve high-quality portrait reenactment. When driving face images using landmarks, it may result in identity mismatch since the landmarks cannot be adapted between different subjects~\cite{DBLP:conf/iccv/ZakharovSBL19}. To alleviate this problem, landmark disentanglement is proposed to isolate geometry and identity~\cite{DBLP:conf/aaai/HaKKSK20,DBLP:journals/corr/abs-2004-12452}. However, without 3D guidance of facial geometry, these 2D image editing methods fail to guarantee consistent quality under large changes of 3D facial attributes.

\subsection{Latent Space Face Manipulation}
The most popular GAN models for face generation, StyleGAN \cite{DBLP:conf/cvpr/KarrasLA19} and its improved version StyleGAN2 \cite{DBLP:conf/cvpr/KarrasLAHLA20}, have demonstrated great success in high-fidelity face image synthesis from one learned latent space.
However, they cannot control the generated results in a semantically controllable and disentangled manner. To overcome this limitation, some recent works \cite{DBLP:conf/cvpr/ShenGTZ20,DBLP:conf/nips/HarkonenHLP20,DBLP:journals/corr/abs-2007-06600} focus on the manipulation of the underlying learned latent space, using the vector arithmetic property observed in \cite{DBLP:conf/nips/MikolovSCCD13}, that is, semantic editing operations can be achieved by first computing a difference vector between two latent vectors and then adding it onto another latent vector. The key to enabling individual control of each attribute is to find the directional vector corresponding to this attribute. InterFaceGAN~\cite{DBLP:conf/cvpr/ShenGTZ20} proposes to learn a binary attribute classifier to identify the separation boundary for each attribute and make the separation boundaries of different attributes as orthogonal as possible. Such attribute classifiers are also used in the very recent work StyleFlow~\cite{DBLP:journals/corr/abs-2008-02401}, which formulates conditional exploration as conditional continuous normalizing flows in the latent space. Both methods rely on a dataset with all the attributes labeled, which requires manual annotation and limits the supported attribute types. In contrast, the noteworthy work GANSpace \cite{
DBLP:conf/nips/HarkonenHLP20} can identify important latent directions for different attributes in an unsupervised fashion. It applies principal component analysis either in latent space or feature space and achieves interpretable control with layer-wise perturbation along with the principal directions. To transfer such controllability to real image editing, the only extra step is to map the real image into the latent space via some GAN inversion methods~\cite{DBLP:conf/iccv/AbdalQW19,DBLP:conf/eccv/ZhuSZZ20}. Broadly speaking, our method also belongs to latent space manipulation, but we provide more fine-grained and rig-like control like~\cite{DBLP:conf/cvpr/TewariEBBSPZT20} by building a bidirectional mapping between the latent space and the 3DMM space.

\subsection{3D Guided Face Manipulation}
3DMM~\cite{DBLP:conf/siggraph/BlanzV99} represents human faces in parametric spaces. It enables face reconstruction from images, provides flexible and explicit control over 3D facial attributes, and further renders the reconstructed faces with illumination models such as spherical harmonics~\cite{DBLP:conf/siggraph/RamamoorthiH01a}.
This model is used in various manipulation applications~\cite{DBLP:journals/tog/DaleSJVMP11,DBLP:journals/cgf/GarridoVSSVPT15,DBLP:journals/tog/ThiesZNVST15,DBLP:journals/cgf/ZollhoferTGBBPS18,DBLP:journals/tog/SuwajanakornSK17}, but they only capture the coarse geometry due to the mostly linear statistical model. Also, computer graphics rendering often leads to non-photorealistic results. Neural rendering techniques are developed to improve the photorealism by using differentiable graphics pipelines~\cite{DBLP:conf/iccv/TewariZK0BPT17,DBLP:journals/tog/KimCTXTNPRZT18,DBLP:journals/tog/KimEZSBRT19}, combining the traditional pipelines with learnable components~\cite{DBLP:journals/tog/ThiesZN19}, or transforming synthetic images into the photorealistic domain~\cite{DBLP:conf/eccv/GecerBKK18,DBLP:journals/tog/FriedTZFSGGJTA19}. These methods can modify facial region realistically, but cannot complete the occluded parts, such as hair, body, and background.
\textcolor{black}{Smith \textit{et al.} \cite{DBLP:conf/cvpr/SmithSDTTE20} improve 3DMM in modeling albedo by using uncalibrated multi-view stereo to compute geometry, warping a template to the raw scanned meshes, and stitching seamless per-vertex diffuse and specular albedo maps. However, this method fails to render diverse photorealistic images, with a limited number of scans. In contrast, Lattas \textit{et al.} \cite{DBLP:conf/cvpr/LattasMGPTGZ20} propose AvatarMe to create photorealistic 3D faces from a single in-the-wild image by training image translation networks to estimate high-quality diffuse and specular albedo using 200 real scans. However, it also suffers from the limited diversity of generated faces and the reconstructed results show blurry high-frequency details.}

The recent research proposes to combine GAN with 3D priors to leverage its capacity in generating high-resolution photorealistic images of human faces~\cite{DBLP:conf/cvpr/GengCT19,DBLP:conf/cvpr/XuYCWDJT20,DBLP:conf/cvpr/DengYCWT20}. The representative work StyleRig~\cite{DBLP:conf/cvpr/TewariEBBSPZT20} provides a face rig-like 3D control over a pre-trained StyleGAN2 by mapping the control space of 3DMM to the latent space of StyleGAN2, but it fails to preserve the visual quality when manipulating real images. The following work PIE~\cite{DBLP:journals/tog/TewariE0BSPZT20} addresses this by embedding real images in the StyleGAN2 latent space via hierarchical optimization. However, the latent space in these methods is entangled, which cannot ensure the consistency of unchanged attributes and backgrounds (\textit{e.g.} hair, clothes). Moreover, such methods cannot be applied to out-of-domain faces because building 3DMM for out-of-domain faces is impractical and expensive. We instead propose a disentangled latent space, which achieves better disentanglement among different face attributes and enables simultaneous control over them and further generalizes the rig-like control to out-of-domain face images via cross-domain adaption.

\begin{figure*}[ht!]
      \centering
         \includegraphics[width=\textwidth]{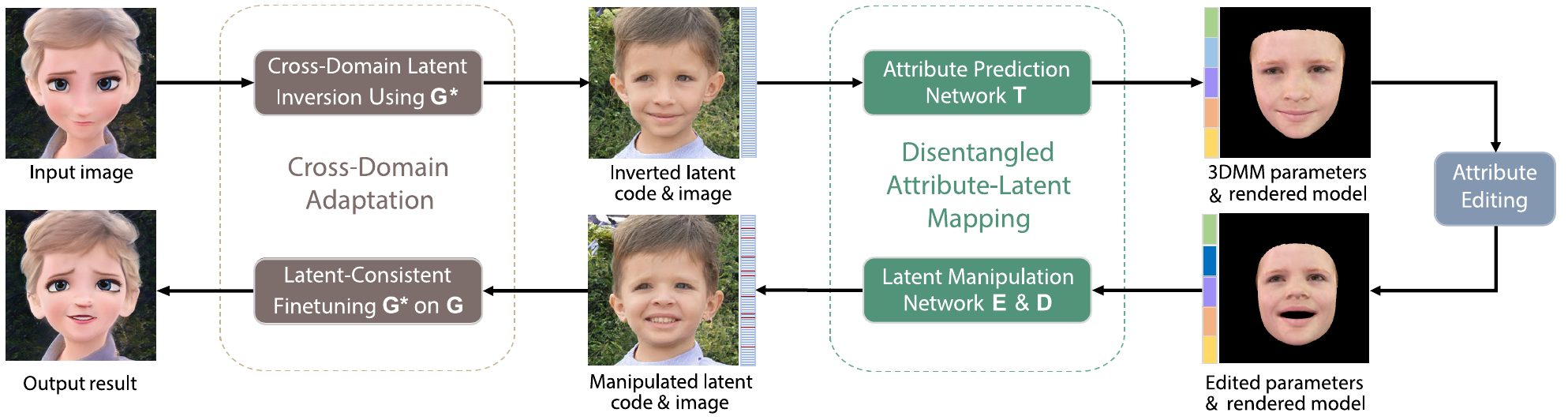}
         \caption{The pipeline of the proposed method consists of two major stages: \textit{cross-domain adaptation} \textcolor{black}{(\S~\ref{sec:domain})} and \textit{disentangled attribute-latent mapping} \textcolor{black}{(\S~\ref{sec:manipulation})}. Given an input out-of-domain face image, the pipeline first reverts it into a latent code in the human face domain \textcolor{black}{(\S~\ref{subsec:inversion})} and then maps the latent code to 3DMM parameters via an attribute prediction network \textcolor{black}{(\S~\ref{subsec:l2a})}. Edits on the pose, expression, illumination, etc., can be directly made to specific 3DMM parameters and then projected to disentangled latent code via a latent manipulation network \textcolor{black}{(\S~\ref{subsec:a2l})}. Finally, the manipulated latent code is fed into a StyleGAN2 finetuned with our latent-consistent finetuning techniques to generate the out-of-domain face image, reflecting the corresponding edits \textcolor{black}{(\S~\ref{subsec:finetuning})}. We edit pose and expression in this example.}
      \label{fig:overview}
\end{figure*}

\section{Background and Overview}
\label{sec:overview}

\subsection{Facial Attribute Parameterization}
\label{subsec:3dmm}

We adopt the widely-used 3D Morphable Model (3DMM)~\cite{DBLP:conf/siggraph/BlanzV99} to parameterize the facial attributes, which provides compact semantic controls over both shape and texture.

In this work, we aim to support all the semantics that 3DMM can offer, \textit{i.e.}, \textit{geometry attributes} of identity $\bm{\alpha}$, expression $\bm{\beta}$, and pose $\bm{T}$; and \textit{appearance attributes} of albedo $\bm{\delta}$ and illumination $\bm{\gamma}$. Specifically, $\bm{\alpha},\bm{\delta}\in\mathbb{R}^{80}$ are the coefficients of morphable bases built through PCA, $\bm{\beta}\in\mathbb{R}^{64}$ are the coefficients corresponding to pre-defined expression bases, $\bm{T}$ is a rigid transformation containing both rotation $\bm{r}\in\textbf{SO}(3)$ and translation $\bm{t}\in\mathbb{R}^3$, and $\bm{\gamma}\in\mathbb{R}^{9\times3}$ are the order-3 spherical harmonics (SH) parameters for RGB channels. Here, we use the expression bases provided by~\cite{DBLP:journals/pami/GuoZCJZ19}, which are built from FaceWarehouse~\cite{DBLP:journals/tvcg/CaoWZTZ14}. \textcolor{black}{This is a common choice that has been widely adopted by many recent works, such as StyleRig \cite{DBLP:conf/cvpr/TewariEBBSPZT20}, Deng \textit{et al.}'s work \cite{DBLP:conf/cvpr/DengYX0JT19} and their improved version \cite{deng2019accurate}. We thus follow their choice on 3DMM bases for a fair comparison.}
We denote this set of control parameters as $\bm{P}=(\bm{\alpha},\bm{\beta},\bm{T},\bm{\delta},\bm{\gamma})$ in the attribute space $\mathcal{P}$.

For the morphable model with $N_v$ vertices, given the average geometry and appearance $\{\overline{\bm{G}},\overline{\bm{A}}\}\in\mathbb{R}^{3\times N_v}$ and basis vectors $\widetilde{\bm{G}}^I,\widetilde{\bm{G}}^E,\widetilde{\bm{A}}$ for identity, expression, and albedo, respectively, we have the interpolated geometry \textcolor{black}{$\bm{G}=\overline{\bm{G}}+\widetilde{\bm{G}}^I\bm{\alpha}+\widetilde{\bm{G}}^E\bm{\beta}$ and appearance $\bm{A}=\overline{\bm{A}}+\widetilde{\bm{A}}\bm{\delta}$}. Ultimately, the final positions $\text{V}(\bm{P})$ and colors $\text{C}(\bm{P})$ of all vertices are evaluated on each individual $i$-th vertex as follows:
\begin{equation}
\textcolor{black}{\text{V}(\bm{P})_i=\bm{G}_i\bm{r}+\bm{t}},
\label{eq:v}
\end{equation}
\begin{equation}
\text{C}(\bm{P})_i=\bm{A}_i\cdot\begin{matrix}\sum_{b=1}^9\end{matrix}\bm{\gamma}_b\cdot\text{H}_b(\bm{n}_i),
\label{eq:c}
\end{equation}
where $\{\bm{G}_i,\bm{A}_i\}\in\mathbb{R}^3$ are the vectors of the $i$-th vertex, $\text{H}_b(\bm{n})\in\mathbb{R}$ is the response of the SH basis function $\text{H}_b$ at normal direction of $\bm{n}$, and $\bm{\gamma}_b\in\mathbb{R}^3$ are the illumination coefficients for the $b$-th SH band.

\subsection{Overview of Our Pipeline}
\label{subsec:overview}

Our cross-domain 3D-guided face manipulation pipeline is illustrated in Fig.~\ref{fig:overview}. Given an input in-the-wild source image, the pipeline first applies domain-consistent latent inversion (\S~\ref{subsec:inversion}) to optimize a latent code that can best reconstruct the image. Then, the pipeline updates the latent code (\S~\ref{sec:manipulation}) to reflect various attribute manipulations controlled by 3DMM parameters (\S~\ref{subsec:3dmm}). This step basically consists of two parts: an attribute prediction network (\S~\ref{subsec:l2a}) that predicts the 3DMM parameters from the latent code and a latent manipulation network (\S~\ref{subsec:a2l}) that projects the parameter manipulation back onto the latent code. Both of them are built upon our reduced style space \textcolor{black}{(\S~\ref{subsec:stylespace})} constructed through our attribute-adaptive layer selection method (\S~\ref{subsec:selection}) to ensure disentangled and precious controls. Finally, our space-consistent domain adaptation approach (\S~\ref{subsec:finetuning}) is adopted to map the manipulated latent code to an arbitrary out-of-domain face domain to achieve cross-domain face editing.

\section{Attribute-Adaptive Latent Decomposition}
\label{sec:latent}

\subsection{Reduced Style Space}
\label{subsec:stylespace}

A pre-trained StyleGAN2~\cite{DBLP:conf/cvpr/KarrasLAHLA20} generator serves as a non-linear function $\text{G}$ that maps the latent feature space $\mathcal{W}^+$ to the image space: $\bm{I}=\text{G}(\bm{w})$, where $\bm{I}$ is the generated image, and $\bm{w}\in\mathbb{R}^{18\times512}$ is the latent vector. The raw $\mathcal{W}^+$ space is a concatenation of $18$ vectors of dimension $512$, where each vector is transformed to channel-wise style parameters~\cite{DBLP:conf/iccv/HuangB17}. The space spanned by these style parameters is referred to as the style space~\cite{DBLP:journals/corr/abs-2011-12799}, denoted as $\mathcal{S}$. For the target generation resolution of $1024\times1024$, a latent code in $\mathcal{S}$ consists of style parameters for $26$ layers with a total dimension of $9088$, in which $17$ layers of dimension $6048$ apply to the feature maps, while the other $9$ tRGB layers takes the remaining dimension of $3040$.

Instead of $\mathcal{W}^+$, we adopt this style space $\mathcal{S}$ as it is demonstrated to be more disentangled than other latent spaces including $\mathcal{W}^+$ on embedding semantic properties~\cite{DBLP:journals/corr/abs-2011-12799}, \textit{i.e.}, changing one attribute tends to be reflected on isolated variations in a few channels of $\mathcal{S}$ without significant overlapping. Taking advantage of this desired property, we propose to further decompose the whole $\mathcal{S}$ space into a certain number of subspaces, with each corresponding to one specific facial attribute. We call this set of subspaces the reduced style space, where each attribute $\bm{P}^i$ is matched to a subspace $\mathcal{S}^i\subseteq\mathcal{S}$. Therefore, given a complete latent code $\bm{w}\in\mathcal{S}$, when manipulating one attribute $\bm{P}^i$ of its corresponding image, only part of the code $\bm{w}^i\in\mathcal{S}^i$ should be altered accordingly. With such an attribute-adaptive reduced space, we have a compact and disentangled attribute embedding in the latent space, which not only facilitates the training performance but also makes it possible to edit one attribute without affecting irrelevant content or manipulate multiple attributes simultaneously without mutual interference.

\begin{figure*}[ht!]
      \centering
      \includegraphics[width=\textwidth]{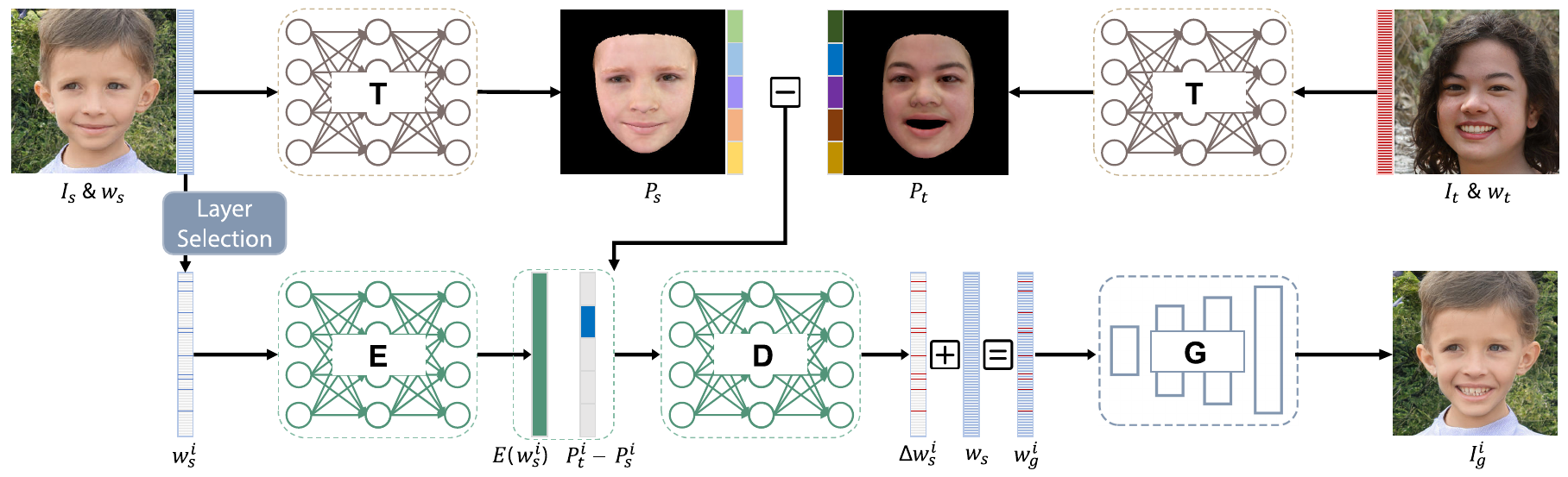}
       \caption{Framework of disentangled face manipulation. Given a source latent code $\bm{w}_s$, to manipulate the face $\bm{I}_s$ encoded by this code, the first step is to interpret it to a set of 3D facial attributes $\bm{P}_s$ through an attribute prediction network \text{T}. And then we take the difference of $i$-th attribute (expression in this example) between $\bm{P}_s$ and the target parameter $\bm{P}_t$, getting $\Delta{\bm{P}^i}=\bm{P}_t^i-\bm{P}_s^i$. $\Delta{\bm{P}^i}$ together with the source code $\bm{w}_s^i$ in subspace $\mathcal{S}^i$ is fed into a latent manipulation network (composed of an encoder \text{E} and a decoder \text{D}), to predict $\Delta{\bm{w}^i}$ for updating source latent code $\bm{w}_s$ and generating the edited result $\bm{I}_g^i$ with a pretrained StyleGAN2 \text{G}. }
      \label{fig:framework}
\end{figure*}

\subsection{Attribute-Adaptive Layer Selection}
\label{subsec:selection}

\setlength{\algomargin}{0pt}
\begin{algorithm}[h]
\footnotesize
\SetAlgoLined
\SetKwInOut{Input}{Input}
\SetKwInOut{Output}{Output}
\SetKwInOut{Define}{Define}
\Input{Pre-trained StyleRig network $\text{S}$ on the $\mathcal{S}$ space;\\Data pairs $\{(\bm{w}_1,\bm{P}_1),(\bm{w}_2,\bm{P}_2),\dots,(\bm{w}_N,\bm{P}_N)\}$;}
\Output{Selected layers $\mathbb{L}^i$ for attribute $\mathcal{P}^i$;}
\Define{$\text{REP}(\bm{P}_a,\bm{P}_b,i)=\{\bm{P}_a^1,\dots,\bm{P}_a^{i-1},\bm{P}_b^i,\bm{P}_a^{i+1},\dots,\bm{P}_a^{|\mathcal{P}|}\}$;\\$\text{COV}(l,c,i)=\frac{1}{N^2}\sum_{a,b}|(\text{S}(\bm{w}_a,\text{REP}(\bm{P}_a,\bm{P}_b,i))-\bm{w}_a)^{(l,c)}||\bm{P}_a-\bm{P}_b|$;}
\BlankLine
$d\gets3$;\\
$\mathbb{L}^i\gets\emptyset$;\\
\While{$|\mathbb{L}^i|<9$}{
    \For{$l\in[1,N_l]$}{
        \For{$c\in[1,N_c^l]$}{
            \lIf{\textup{$\text{COV}(l,c,i)>d$}}{\textup{$n^{(l,c)}\gets1$}}
            \lElse{\textup{$n^{(l,c)}\gets0$}}
        }
    \lIf{\textup{$\sum_{c=1}^{N_c^l}n^{(l,c)}>0.25*N_c^l$}}{\textup{$\mathbb{L}^i\gets\mathbb{L}^i\cup\{l\}$}}
    }
    $d\gets0.95*d$;
}
\caption{Attribute-Adaptive Layer Selection Algorithm}
\label{alg:select}
\end{algorithm}

To faithfully select latent layers for each attribute, our intuition is to have an \textit{agent} that maps the variation on each attribute to different layers in the space. We select those layers with strong responses since they tend to have strong correlations with the semantics controlled by that attribute.

The best choice for such a mapping agent should be driven by tasks relevant to encoding the 3DMM attributes to the latent space. Hence, we use StyleRig implemented on the style space as an agent, which demonstrates high-quality attribute-to-latent encoding. 
Here we denote the StyleRig network simply as a function $w_g=\text{S}(w_s,\bm{P}_t)$, which transforms the source code $w_s$ to $w_g$ controlled by the target attribute parameters $\bm{P}_t$.

We propose the selection algorithm as detailed in Alg.~\ref{alg:select}, which eventually selects a set of layers $\mathbb{L}^i$ for each attribute $\mathcal{P}^i$ that construct a corresponding subspace $\mathcal{S}^i$. Specifically, we measure the correlation between $i$-th attribute and the $c$-th channel of the $l$-th layer through the covariance $\text{COV}(l,c,i)$ between the latent values $\bm{w}^{(l,c)}$ and the attribute parameter $\bm{P}^i$. Specifically, given one random attribute-latent pair $(\bm{w},\bm{P})$, we randomly change $\bm{P}^i$ and calculate the corresponding covariance $\text{COV}(l,c,i)$ for all channels of $l$-th layer. A layer is selected if more than a quarter of its channels have the covariance larger than a threshold $d$. And this threshold is gradually decreased in case too few layers are selected. In our experiments, the same threshold is used for every parameter/layer.
Although our algorithm is not sensitive to this threshold, a lower value will lead to a faster convergence without influencing results. We set it to 3 empirically.

\section{Disentangled Face Manipulation}
\label{sec:manipulation}

At the core of our face manipulation framework is a mutual mapping structure between the \textit{latent space} $\mathcal{S}$ and the \textit{attribute space} $\mathcal{P}$.
The framework is illustrated in Fig.~\ref{fig:framework}.

\subsection{Attribute Prediction Network}
\label{subsec:l2a}

Given a source latent code $\bm{w}_s\in\mathcal{S}$ as the input to our framework, to manipulate the face encoded by this code, the first step is to interpret it to a set of 3D facial attributes $\bm{P}_s$ that are understandable and controllable. We achieve this through an \textit{attribute prediction network} $\bm{P}=\text{T}(\bm{w})$, which regresses 3DMM parameters $\bm{P}_s$ out of $\bm{w}_s$ to best fit the subject face encoded in the source code. This translation network is also applied to the target latent code $\bm{w}_t$ in the reference-based manipulation mode.

\subsection{Latent Manipulation Network}
\label{subsec:a2l}

After the source attribute parameters $\bm{P}_s$ are predicted with the attribute prediction network, user manipulation is performed in the attribute space, which either changes the parameters from the source $\bm{P}_s$ to the target $\bm{P}_t$ by incrementally editing one or some of the attributes (edit-based manipulation) or entirely transferring from a reference (reference-based manipulation). The purpose of this attribute-to-latent space translation is then to update the latent code to inject the desired manipulation operations.

Thanks to the vector arithmetic properties~\cite{DBLP:journals/corr/RadfordMC15} of the latent space embedding, we can potentially update the latent code via projecting the attribute changes to a linear displacement vector, of which the unit direction controls the identified editing semantics and the magnitude represents the manipulation intensity. This attribute-to-latent translation can be generally formulated as a mapping function $\Delta\bm{w}=\text{F}(\bm{w}_s,\bm{P}_t)$, 
and the updated latent code to be $\bm{w}_g=\bm{w}_s+\Delta\bm{w}$.

The RigNet proposed in StyleRig~\cite{DBLP:conf/cvpr/TewariEBBSPZT20} also falls into this formulation. Specifically, it contains an encoding structure $\text{E}'$ that transforms the source code to feature vectors, and a decoding structure $\text{D}'$ that takes both the feature vector and the target parameters to generate the final latent displacement vector: $\Delta\bm{w}=\text{F}'(\bm{w}_s,\bm{P}_t)=\text{D}'(\text{E}'(\bm{w}_s),\bm{P}_t)$, where $\bm{w}\in\mathcal{W}^+$. However, despite its huge success in achieving promising parametric 3D controls, there exist three types of \textit{entanglements} that prevent more precise and flexible manipulation:
\begin{itemize}
    \item The source latent code $\bm{w}_s$ and the absolute target parameters $\bm{P}_t$ partially share the identity information of the subject;
    \item Correlations between different parameters in $\bm{P}_t$ prevent isolated manipulation through the translation networks;
    \item The $W^+$ space is not well disentangled regarding the semantics; editing one attribute could potentially affect the others.
\end{itemize}

In light of these issues, we adopt a \textit{latent manipulation network} to achieve disentangled attribute-to-latent mapping, as shown in Fig.~\ref{fig:framework}. By using our reduced style space that decomposes $\mathcal{S}$ into attribute-adaptive subspaces $\{\mathcal{S}^1,\mathcal{S}^2,\dots,\mathcal{S}^{|\mathcal{P}|}\}$, instead of jointly manipulating the entire target latent space $\mathcal{W}^+$, our framework consists of independent small translation networks $\{\text{F}^1,\text{F}^2,\dots,\text{F}^{|\mathcal{P}|}\}$, with each handles a specific target subspace of $\mathcal{S}^i$ corresponding to an attribute $\bm{P}^i$ decided by our reduced style space. In addition, rather than manipulating the latent code on the absolute target parameters $\bm{P}_t$, we feed the relative changes of parametric edits $\Delta\bm{P}=\bm{P}_t-\bm{P}_s$ to enforce the focus on the manipulation itself instead of the original identity. Our final attribute-to-latent translation framework is formulated as:
\begin{equation}
\Delta\bm{w}^i=\text{F}^i(\bm{w}^i_s,\bm{P}^i_t-\bm{P}^i_s)=\text{D}^i(\text{E}^i(\bm{w}^i_s),\bm{P}^i_t-\bm{P}^i_s),
\label{eq:w1}
\end{equation}
\begin{equation}
\bm{w}^i_g=\bm{w}_s+\Delta\bm{w}^i.
\label{eq:w2}
\end{equation}
Here, each network $\text{F}^i$ consists of an encoder $\text{E}^i$ and a decoder $\text{D}^i$.

\subsection{Training Strategy}
\label{subsec:training}

Data Preparation:
The training data we use to train the system generally consists of tuples $(\bm{w},\bm{I},\bm{P})$, in which $\bm{w}\in\mathcal{S}$ is the latent code in the style space, $\bm{I}=\text{G}(\bm{w})$ is the image generated with StyleGAN2 generator $\text{G}$ corresponding to the code $\bm{w}$, and $\bm{P}\in\mathcal{P}$ is the 3DMM parameters of the subject in $\bm{I}$.

To prepare such a tuple, we first independently sample $5$ latent vectors of size $512$ in the $\mathcal{W}$ space from a normal distribution. Then, to construct the final code $\bm{w}'\in\mathcal{W}^+$ of dimension $18\times512$, which is essentially a concatenation of $18$ size-$512$ vectors, we randomly select one from the $5$ latent vectors for $18$ times in a row and stack them together.
After that, we convert $\bm{w}'$ to $\bm{w}\in\mathcal{S}$ and generate the face image $\bm{I}=\text{G}(\bm{w})$. This technique is inspired by the mixing regularizer~\cite{DBLP:conf/cvpr/KarrasLAHLA20}, which helps prevent the generator from assuming that the adjacent latent codes are correlated and improve the quality of the synthesized images.
Finally, we predict the 3D parameters $\bm{P}$ from the image $\bm{I}$ using the off-the-shelf 3D face reconstruction method~\cite{DBLP:conf/cvpr/DengYX0JT19}.

Pre-Training for Attribute Prediction:
To avoid over-complicating the main training phase, we pre-train the attribute prediction network $\text{T}$ in a supervised manner with ground truth pairs $(\bm{w},\bm{P})$. Considering the varying contribution to the final 3D model of different parameters due to their distinct nature and strong correlation, we assign a weight for each individual parameter which is dynamically adjusted during the training. Specifically, we adopt the weighted parameter distance cost (WPDC) loss~\cite{DBLP:conf/cvpr/ZhuLLSL16}:
\begin{equation}
\mathcal{L}_{wpdc}=\left\|e\cdot\left(\bm{P}-\text{T}(\bm{w})\right)\right\|_2^2,
\label{eq:wpdc}
\end{equation}
where the parameter weight $e$ is defined as the shape/color deviation caused solely by this parameter at the current value $\text{T}(\bm{w})$ if all other parameters are replaced with the groud truth $\bm{P}$:
\begin{equation}
\begin{split}
e_i=&\left\|\text{V}(\text{REP}(\bm{P},\text{T}(\bm{w}),i))-\text{V}(\bm{P})\right\|_2\\
+&\left\|\text{C}(\text{REP}(\bm{P},\text{T}(\bm{w}),i))-\text{C}(\bm{P})\right\|_2.
\end{split}
\label{eq:e}
\end{equation}
Here $\text{REP}$ is the replacement function that replaces the $i$-th element of $\bm{P}_a$ with $\bm{P}_b$, as defined in Alg.~\ref{alg:select}.

Self-Supervised Training for Latent Manipulation:
As ground truth pairs are generally unavailable for in-the-wild images regarding most semantic manipulation operations, we adopt a self-supervised training solution to train the latent manipulation network. 

Following StyleRig~\cite{DBLP:conf/cvpr/TewariEBBSPZT20}, training is performed on latent code pairs $\{\bm{w}_s,\bm{w}_t\}\in\mathcal{S}$ together with their tuples $(\bm{w}_s,\bm{I}_s,\bm{P}_s)$ and $(\bm{w}_t,\bm{I}_t,\bm{P}_t)$. The basic idea is to let the network manipulate the source code $\bm{w}_s$ by replacing one single attribute of $\bm{P}^i_s$ to $\bm{P}^i_t$. Through the networks $\{\text{T},\text{E},\text{D}\}$, the generated latent code is calculated as:
\begin{equation}
\textcolor{black}{\bm{w}_g=\bm{w}_s+\text{D}(\text{E}(\bm{w}_s),\text{REP}(\text{T}(\bm{w}_s),\text{T}(\bm{w}_t),i)-\text{T}(\bm{w}_s)),}
\label{eq:wg}
\end{equation}
with its parameters as $\bm{P}_g=\text{T}(\bm{w}_g)$, 
where $\text{REP}$ is the replacement function defined in Alg.~\ref{alg:select}. Given $\bm{w}_g$, we can enforce various self-supervision to ensure that the resultant image $\bm{I}_g=\text{G}(\bm{w}_g)$ conforms to the manipulation $\bm{P}^i_s\to\bm{P}^i_t$, while keeping other parameters $\bm{P}^j_s,j\neq i$ unchanged. For simplicity, here we define two terms that are important to the self-supervisions: $\bm{P}_{tg}=\text{REP}(\bm{P}_t,\bm{P}_g,i)$ and $\bm{P}_{gs}=\text{REP}(\bm{P}_g,\bm{P}_s,i)$, which should be as close as possible to the target and source parameters, respectively, to measure the accuracy and the quality of disentanglement of the manipulation.

First of all, we utilize the differentiable renderer to measure photometric similarities in the image space and back-propagate them to the attribute space. This rendering loss is defined as:
\begin{equation}
\mathcal{L}_{render}=\|\text{R}(\bm{P}_{tg})-\text{M}(\bm{P}_{t})\cdot\bm{I}_t\|_2^2+\|\text{R}(\bm{P}_{gs})-\text{M}(\bm{P}_{s})\cdot\bm{I}_s\|_2^2,
\label{eq:render}
\end{equation}
where $\text{R}(\bm{P})$ is the differentiably rendered face image from parameters $\bm{P}$, with its corresponding occupancy mask $\text{M}(\bm{P})$, \textit{i.e.}, 2D regions covered by the projected morphable model. The dimensions of these images are made equal to $\bm{I}_s$ and $\bm{I}_t$.

By using sparse landmarks that are pre-labeled on the mesh, we also adopt the landmark loss:
\begin{equation}
\begin{split}
\mathcal{L}_{land}=&\|\Pi(\text{L}(\text{V}(\bm{P}_{tg})))-\Pi(\text{L}(\text{V}(\bm{P}_t)))\|_2^2\\
+&\|\Pi(\text{L}(\text{V}(\bm{P}_{gs})))-\Pi(\text{L}(\text{V}(\bm{P}_s)))\|_2^2,
\end{split}
\label{eq:land}
\end{equation}
where the function $\text{L}(\bm{G})\in\mathbb{R}^{N_l\times3}$ samples the 3D landmark positions from mesh vertices $\bm{G}$ and the function $\Pi(\bm{L})\in\mathbb{R}^{N_l\times2}$ projects landmarks $\bm{L}$ to 2D positions on the image plane with the weak-perspective camera model. We use $N_l=68$ in our experiments. \textcolor{black}{Landmark positions are 
provided by \cite{DBLP:conf/cvpr/DengYX0JT19}. }

Similarly, we propose the contour loss to further improve the consistency of face shape:
\begin{equation}
\begin{split}
\mathcal{L}_{cont}=&\|\text{L}'(\text{V}(\bm{P}_{tg}))-\text{L}'(\text{V}(\bm{P}_t))\|_2^2\\
+&\|\text{L}'(\text{V}(\bm{P}_{gs}))-\text{L}'(\text{V}(\bm{P}_s))\|_2^2,
\end{split}
\label{eq:cont}
\end{equation}
where $\text{L}'(\bm{G})\in\mathbb{R}^{N_c\times3}$ samples only a subset of landmarks that are related to the face outer contour ($N_c$=17). Different from the 2D landmark loss $\mathcal{L}_{land}$ (Eq.~\ref{eq:land}), this contour loss is defined on 3D positions without transformation-dependent projection, which avoids the pose bias due to inaccurate rotation.

\begin{equation}
\mathcal{L}=\lambda_r\mathcal{L}_{render}+\lambda_l\mathcal{L}_{land}+\lambda_c\mathcal{L}_{cont}.
\label{eq:l}
\end{equation}

\section{Cross-Domain Adaptation}
\label{sec:domain}

Using our in-domain latent-attribute mapping pipeline, we are able to perform disentangled manipulation of human face images. In this section, we introduce how we adapt this 3D-guided face manipulation framework to handle cross-domain faces, including non-photorealistic paintings, cartoon portraits, animals, etc.

\subsection{Latent-Consistent Finetuning}
\label{subsec:finetuning}

As introduced in \S~\ref{sec:manipulation}, based on the StyleGAN2 generator $\text{G}$ that is pre-trained on the real-face domain $\mathcal{A}$, we edit latent code in the latent space $\mathcal{S}$ via its mutual mapping with the facial attribute space $\mathcal{P}$ parameterized by human 3DMM. To make the best of the well-defined parameterization on human faces and to maintain the consistency of our disentangled latent embedding, we propose to adapt the in-domain latent space to out-domain by performing domain-consistent finetuning of the generator to ensure that the cross-domain latent code could be effectively mapped to each domain-specific generator. The key for the domain adaptation is to finetune StyleGAN2 with proper regularization to ensure the alignment of latent spaces. Different from the existing methods~\cite{DBLP:arxiv/huang2020unsupervised,DBLP:toonify} that align the $\mathcal{W}$ space by \textcolor{black}{swapping or fixing} low-level convolutional layers, we innovatively align the style space by fixing the affine transformation layers, the weights converting the latent code from $\mathcal{W}^+$ to $\mathcal{S}$. This design is tailored for our disentangled mapping in the style space so as to make the editings consistent across different domains.

Specifically, for a new face domain $\mathcal{B}$, we train a generator $\text{G}^*$ on $\mathcal{B}$ which shares a same latent embedding as $\text{G}$. Given a source image $\bm{I}_s\in\mathcal{B}$, we apply cross-domain latent inversion to get latent code $\bm{w}_s\in\mathcal{S}$ so that $\text{G}^*(\bm{w}_s)\sim\bm{I}_s$. With $\bm{w}_s$, in-domain face manipulation is used to update the latent code to $\bm{w}_g$ and produce the final image $\bm{I}_g=\text{G}^*(\bm{w}_g)$ in domain $\mathcal{B}$.

We sample training images from the target domain $\mathcal{B}$ and use them to finetune the original real-face StyleGAN2 generator $\text{G}$ to get the domain-specific generator $\text{G}^*$ for domain $\mathcal{B}$. However, straightforward finetuning which updates all network parameters would distort the underlying latent space. To keep the latent consistency across all generators, we freeze the network layers relevant to latent embedding. To be specific, since we use the style space $\mathcal{S}$, all style convolution layers and tRGB layers, \textit{i.e.}, the affine transformation layers, are fixed during the finetuning. Besides that, we use exactly the same loss functions and hyper-parameters as in StyleGAN2. Eventually, we have $\text{G}$ and $\text{G}^*$ that share the same latent space $\mathcal{S}$. A manipulated latent code $\bm{w}_g$, conditioned on human-face 3DMM attributes, could be mapped to different domains with the same controlled semantics. \textcolor{black}{For this latent-consistent finetuning, we inherit the learning rate and hyperparameters of StyleGAN2 \cite{DBLP:conf/cvpr/KarrasLAHLA20}.}

\textcolor{black}{Regarding the latent-consistent finetuning, the major difference between our method and AgileGAN \cite{DBLP:journals/tog/SongLLMLZC21} is that we innovatively align the style space by fixing the affine transformation layers and finetuning the remaining layers of StyleGAN2 \cite{DBLP:conf/cvpr/KarrasLAHLA20}, while AgileGAN aligns the \(\mathcal{W}+\) space by directly finetuning all layers of StyleGAN2. Our design is tailored for the target of a disentangled mapping in the style space, and it is shown to achieve better alignment with consistent editings.}

\subsection{Cross-Domain Latent Inversion}
\label{subsec:inversion}

To complete the pipeline that manipulates an in-the-wild image from an arbitrary face domain, the last piece of the puzzle we need is a cross-domain latent inversion component $\text{I}$ that can robustly embed an in-the-wild image from the face domain $\mathcal{B}$ into our latent space $\mathcal{S}$: $\bm{w}_s=\text{I}(\bm{I}_s),\bm{w}_s\in\mathcal{S},\bm{I}_s\in\mathcal{B}$.

We take an optimization-based approach with perceptual and MSE loss between the original and the reconstructed image:
\begin{equation}
\bm{w}'_s=\arg\min_{\bm{w}'}(\|\bm{I}_s-\text{G}^*(\bm{w}')\|_2^2+\begin{matrix}\sum_{l=1}^L\end{matrix}\|\Psi_l(\bm{I}_s)-\Psi_l(\text{G}^*(\bm{w}'))\|_1),
\label{eq:inverse}
\end{equation}
where $\Psi_l(\bm{I})$ computes the activation feature map of image $\bm{I}$ at the $l$-th selected layer of the VGG-19 network pre-trained on ImageNet. Here $\bm{w}'\in\mathcal{W}^+$. We transform it to the $\mathcal{S}$ space using the style convolutional layers $\text{S}$ as $\bm{w}_s=\text{S}(\bm{w}'_s)$.
\textcolor{black}{In this proposed inversion method, we adopt Adam optimizer with 2000 iterations for each inversion task.}.

\section{Experiments}

\subsection{Implementation Details}

\textbf{Training parameters.}
We train both the attribute prediction and the latent manipulation networks using Adam optimizer for $60$ epochs with batch size $128$. The initial learning rate is $0.01$ and decayed by $0.1$ every $10$ epochs. For the latent manipulation network, the weights of loss terms are set as $\lambda_r=1$, $\lambda_l=0.001$, and $\lambda_c=0.01$, respectively. \textcolor{black}{We configure these weight coefficients of loss terms through a coarse grid search manner with values in [0.001,0.01,0.1,1.0,10].} 
To train the domain-specific StyleGAN2 generator $\text{G}^*$ for each domain, we finetune the pretrained StyleGAN2 for $32,000$ iterations with batch size $16$ on each dataset using the same learning rate scheduler but a lower learning rate of $0.002$.

\noindent \textbf{Network architecture.}
The attribute prediction network $\text{T}$ consists of $5$ MLP layers with $4$ ELU activations after each intermediate layer. The intermediate layers have dimensions of $4096$, $2048$, $1024$, and $512$, while the output dimension is $257$. For the latent manipulation network, each encoder $\text{E}$ consists of $N$ MLP layers, with $N$ set to $9$, $4$, $7$, $9$, and $4$ for expression, pose, albedo, illumination, and shape, respectively. And each decoder $\text{D}$ consists of $3$ MLP layers. All the intermediate layers have dimensions of $512$. The input to $\text{E}$ is a latent vector in the reduced style space, and its output dimension is $32$. The input to $\text{D}$ is the concatenation of the output of $\text{E}$ and $\bm{P}_t-\bm{P}_s$, and its output is $\Delta \bm{w}$.
Uniform weight initialization is used for the networks by default. 
Besides, we use Pytorch3D as the differential renderer with weak perspective projection, following StyleRig. The model is trained on an NVIDIA 2080 Ti GPU and converges in about 16 hours. The inference time is about 0.78s on an Intel i7-9700 CPU platform, thus allowing interactive editing, as shown in the supplemental video demo.


\noindent \textbf{Out-of-domain datasets.} We use five out-of-domain face datasets to demonstrate the cross-domain face manipulation effects, including Ukiyo-e Face~\cite{pinkney2020ukiyoe}, AFHQ Dog~\cite{DBLP:conf/cvpr/ChoiUYH20}, WikiArt Dataset~\footnote{https://github.com/cs-chan/ArtGAN/tree/master/WikiArt\%20Dataset}, Danbooru2018~\cite{danbooru2020}, and Disney Face involving $400$ online images of Disney cartoon characters collected by ourselves. We use 3906, 3552, 750, 7500, and 320 images to construct Ukiyo-e Face, AFHQ Dog, WikiArt, Danbooru2018, and Disney Face datasets respectively.

\begin{table}[t]
\caption{Evaluation on the disentanglement quality of attribute manipulation, measured by our disentanglement metric, using different spaces ($\mathcal{W}^+$, reduced $\mathcal{W}^+$, $\mathcal{S}$, and reduced $\mathcal{S}$) for the attribute prediction network and different inputs ($\bm{P}$, $\Delta\bm{P}$) for the latent manipulation network. Lower is better.}
\label{tbl:ab_disentangle}
\tabcolsep=1.4pt
\begin{tabular}{C{70pt}|C{25pt}C{25pt}C{25pt}C{25pt}C{25pt}C{25pt}}
\toprule
Method                 & shape & exp. & illum. & pose & albedo & avg. \\ 
\toprule
$\mathcal{W}^+$ \& $\bm{P}$ & 75.69 & 80.76 & 72.79 & 83.64 & 70.42  & 76.66 \\
$\mathcal{S}$ \& $\bm{P}$ & 73.93 & 78.82  &  69.59  & 69.39  & 69.92  & 72.33 \\
$\mathcal{S}$ \& $\Delta \bm{P}$ & 68.09 & 65.48 & 69.09 & 54.73 & 68.92 & 65.26 \\
reduced $\mathcal{W}^+$ \& $\bm{P}$ & 74.61 & 80.69 & 71.99 & 77.06 & 69.32  & 74.73 \\
reduced $\mathcal{S}$ \& $\bm{P}$& 72.55 & 73.38 & 69.15 & 63.35 & 69.26  & 69.54 \\
reduced $\mathcal{S}$ \& $\Delta \bm{P}$ & \textbf{59.45} & \textbf{60.87} & \textbf{66.29} & \textbf{42.75}  & \textbf{67.42}  & \textbf{59.36} \\
\bottomrule
\end{tabular}
\end{table}


\begin{figure*}[t]
      \centering
      \setlength{\tabcolsep}{0\linewidth}
      \begin{tabular}{C{0.105\linewidth}C{0.02625\linewidth}C{0.105\linewidth}C{0.105\linewidth}C{0.105\linewidth}C{0.105\linewidth}C{0.02625\linewidth}C{0.105\linewidth}C{0.105\linewidth}C{0.105\linewidth}C{0.105\linewidth}}
      \multicolumn{11}{c}{\includegraphics[width=0.9975\textwidth]{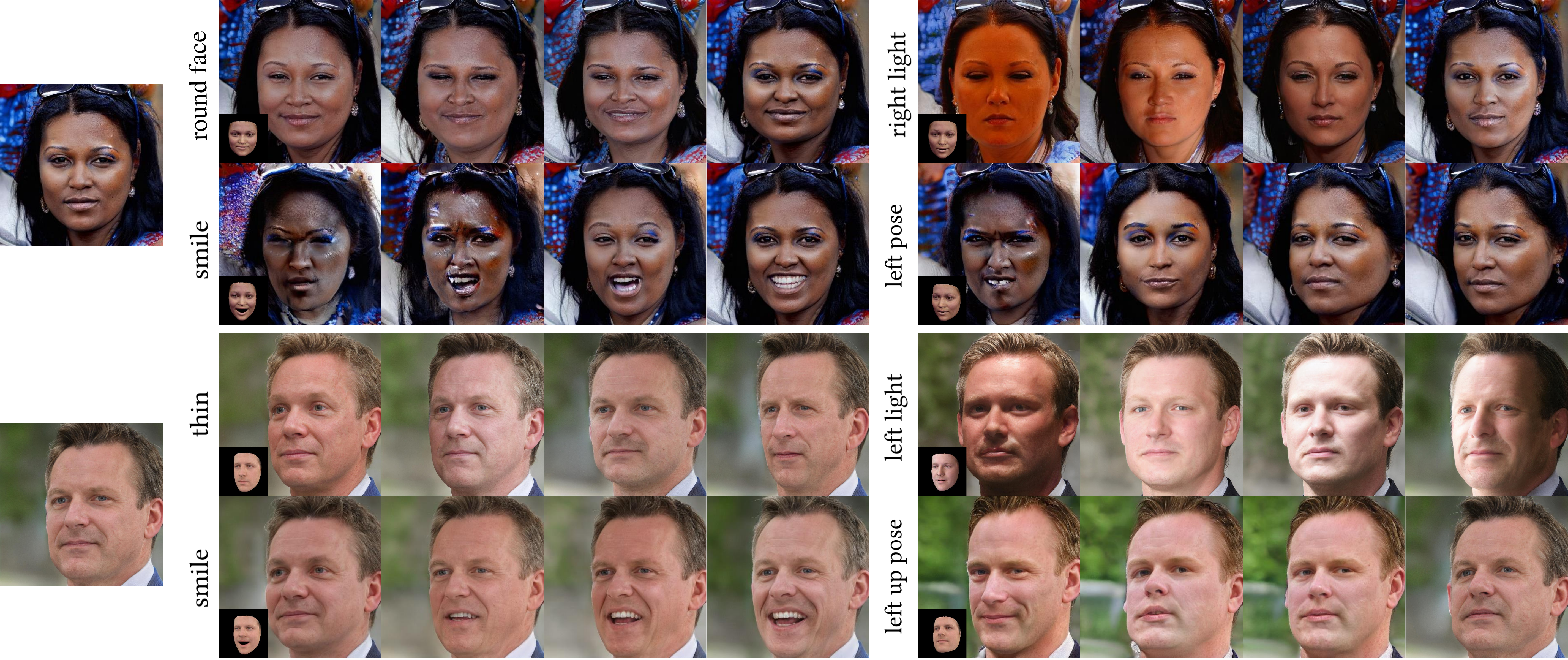}}\\
      original&&(a)&(b)&(c)&(d)&&(a)& (b)& (c)& (d)\\
      \end{tabular}

      \caption{Visual results of different spaces for the attribute prediction network and different inputs for the latent manipulation network. (a): $\mathcal{W}^+$ \& $\bm{P}$. (b): reduced $\mathcal{W}^+$ \& $\bm{P}$. (c): reduced $\mathcal{S}$ \& $\bm{P}$. (d): reduced $\mathcal{S}$ \& $\Delta \bm{P}$.
      With our reduced style space and the relative parametric changes $\Delta\bm{P}$, our results (last column of each case) achieve the best attribute disentanglement quality. Each edit shares the same 3DMM model, shown as the inset in (a).}
      \label{fig:ab_images}
\end{figure*}

\begin{figure}[t]
      \centering
      \setlength{\tabcolsep}{0\linewidth}
      \begin{tabular}{C{0.242\linewidth}C{0.03025\linewidth}C{0.242\linewidth}C{0.242\linewidth}C{0.242\linewidth}}
      \multicolumn{5}{c}{\includegraphics[width=\linewidth]{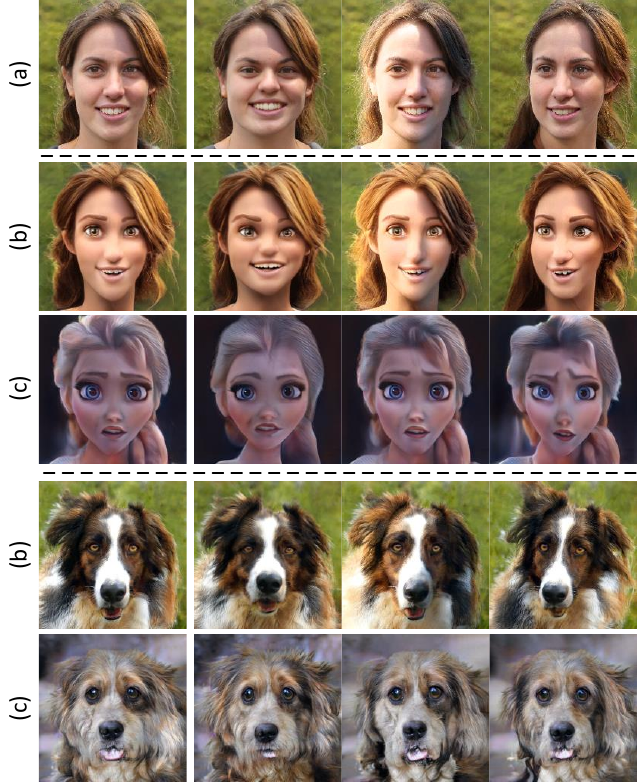}}\\
      \qquad original&& \quad shape& \quad illum.&pose\\
      \end{tabular}
      
      \caption{Ablation study for latent-consistent finetuning. (a): Real human faces. (b): Results with our latent-consistent finetuning. (c): Results without the latent-consistent finetuning. Our method can learn well-aligned latent spaces between $\text{G}$ and $\text{G}^*$, while the generated results without our latent-consistent finetuning show noticeable semantic inconsistency. }
      \label{fig:latent_consist}
\end{figure}

\begin{figure*}[t]
      \centering
      \setlength{\tabcolsep}{0\linewidth}
      \begin{tabular}{C{0.106\linewidth}C{0.01325\linewidth}C{0.106\linewidth}C{0.106\linewidth}C{0.01325\linewidth}C{0.106\linewidth}C{0.106\linewidth}C{0.106\linewidth}C{0.01325\linewidth}C{0.106\linewidth}C{0.106\linewidth}C{0.106\linewidth}}
      &&\multicolumn{2}{c}{shape}&&\multicolumn{3}{c}{turn right}&&\multicolumn{3}{c}{open mouth}\\
      \multicolumn{12}{c}{\includegraphics[width=0.99375\linewidth]{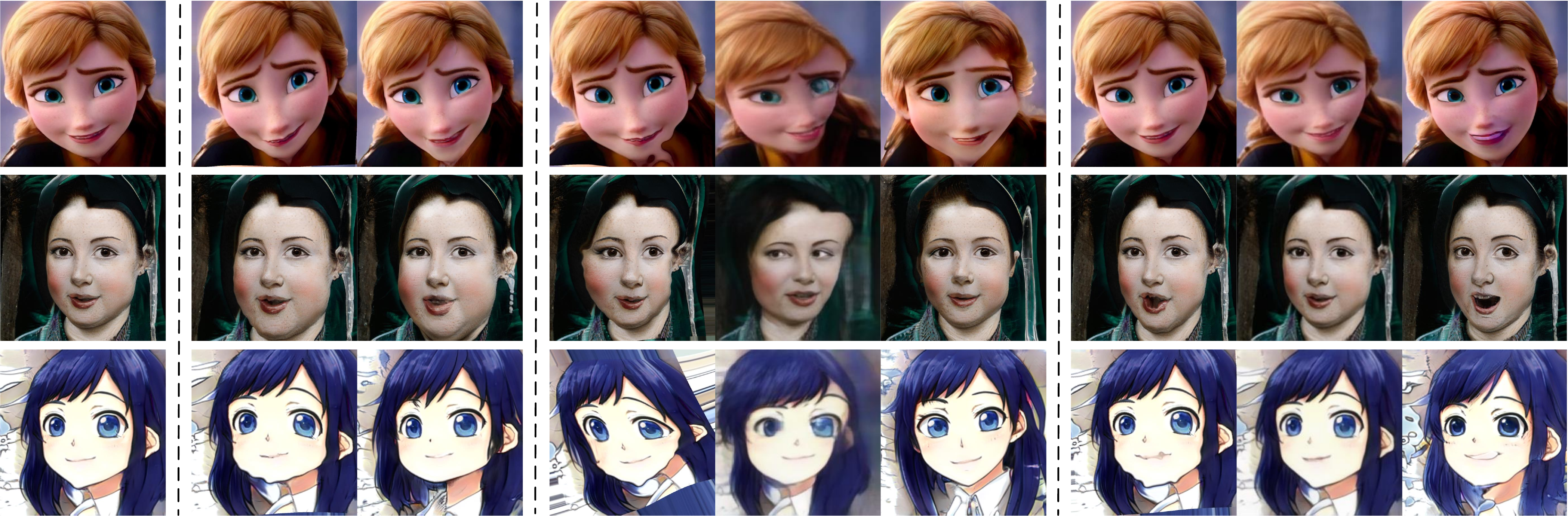}}\\
   original&&MLS&ours&&MLS&First Order&ours&&MLS&First Order&ours\\
      \end{tabular}
      \caption{Comparison results with 2D warping based methods MLS and First Order. Since First Order does not support shape change, we do not include it in the shape editing results. It can be seen that our method achieves better editing results than both MLS and First Order, which either cause noticeable artifacts or fail to achieve the target editing effect (\textit{e.g.} ``Open month'' for First Order).}
      \label{fig:cmp_warp}
\end{figure*}

\begin{figure*}[ht!]
      \centering
      \setlength{\tabcolsep}{0\linewidth}
      \begin{tabular}{C{0.086\linewidth}C{0.01075\linewidth}C{0.086\linewidth}C{0.086\linewidth}C{0.01075\linewidth}C{0.086\linewidth}C{0.104\linewidth}C{0.068\linewidth}C{0.01075\linewidth}C{0.086\linewidth}C{0.086\linewidth}C{0.01075\linewidth}C{0.086\linewidth}C{0.104\linewidth}C{0.068\linewidth}}
      &&\multicolumn{2}{c}{round face}&&\multicolumn{3}{c}{open mouth}&&\multicolumn{2}{c}{left light}&&\multicolumn{3}{c}{turn right}\\
      \multicolumn{15}{c}{\includegraphics[width=0.989\linewidth]{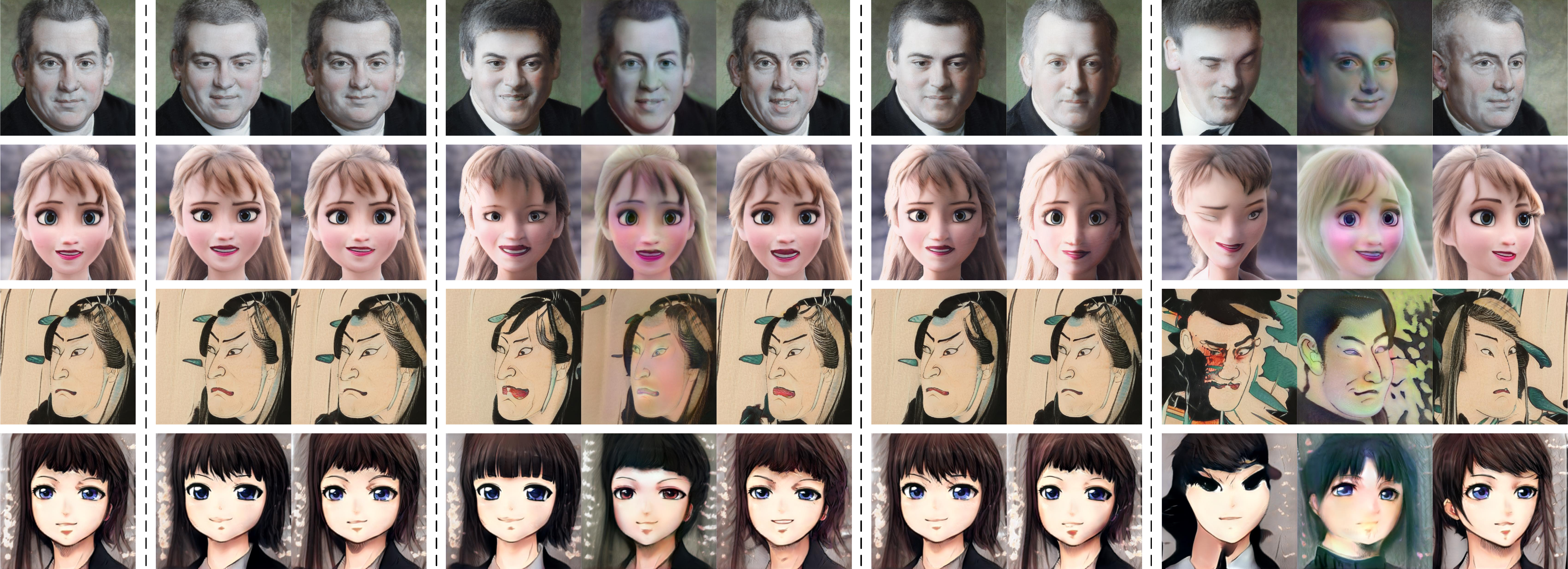}}\\
      original&&GANS.&ours&&GANS.&InterFace.&ours&&GANS.&ours&&GANS.&InterFace.&ours\\
      \end{tabular}
      \caption{Comparison results with GAN latent space manipulation methods: GANSpace~(GANS.) and InterFaceGAN~(Interface.). Since InterFaceGAN does not support shape change and illumination change, we do not provide its results for the first and third case.}
      \label{figure:cmpgan}
\end{figure*}


\subsection{Ablation Study for Disentanglement}
Disentangling different face attributes in the editing process is crucial for enabling the fine-grained controllability of each attribute and guaranteeing the final editing quality. To improve the disentanglement, our method has two key designs: using the reduced style space where one specific facial attribute has one specific subspace rather than the $\mathcal{W}^+$ space used in \cite{DBLP:conf/cvpr/TewariEBBSPZT20}, and using the relative changes of the parametric edits $\Delta \bm{P}$ rather than the absolute target parameters $\bm{P}_t$ in the latent manipulation network.

To quantitatively evaluate the disentanglement for each facial attribute, we randomly sample a set of source latent codes $\{\bm{w}^i_s\}_{i=1}^n$ from the latent space $\mathcal{S}$ and measure the average 3DMM parameter differences before and after editing along the dimension of each attribute respectively. More specifically, for each latent code $\bm{w}^i_s$, we first map it into the corresponding 3DMM parameter $\bm{P}^i_s=\text{T}(\bm{w}^i_s)$ via the attribute prediction network, and then we do the editing in the 3DMM space and get the target edited 3DMM parameter $\bm{P}^i_t$. Next, we map $\bm{P}^i_t$ back to the latent space via the latent manipulation network and get the edited latent code ${\bm{w}}^i_g$. Finally, ${\bm{w}}^i_g$ is mapped to the 3DMM space to get the 3DMM parameter $\bm{P}^i_g$ for the actually edited face. The average L1 distance between ${\bm{P}}^i_g$ and $\bm{P}^i_t$ along the attribute-specific dimension serves as the disentanglement metric for each attribute.
Note that the metric values of different facial attributes cannot be compared directly due to the large disparities of their ranges and variations between attributes. Therefore, we normalize the metrics per attribute by median and standard deviation, and list normalized metrics with different settings in \Tref{tbl:ab_disentangle}.

Benefits of the Reduced Style Space: To demonstrate the advantage of the proposed reduced style space over the $\mathcal{W}^+$ space used in \cite{DBLP:conf/cvpr/TewariEBBSPZT20}, we keep using $\bm{P}$ (not $\Delta\bm{P}$) and replace the reduced style space in our framework with the original $\mathcal{W}^+$ space where different attributes are not disentangled, and the variant ``reduced $\mathcal{W}^+$ space" where each attribute is disentangled and has its own subspace described in \S~\ref{subsec:stylespace}. As shown in \Tref{tbl:ab_disentangle}, using independent reduced subspaces for each attribute can improve disentanglement by reducing the metric from $76.66\%$ to $74.73\%$. By using the reduced style space rather than $\mathcal{W}^+$ Space, the disentanglement metric can be further reduced to $69.54\%$,
which also shows better performance than $72.33\%$ of the style space.
We think that the underlying reason for improvements mainly comes from two aspects: 1) Using an independent subspace for each attribute can not only avoid the interference from other attributes but also reduce the learning difficulty on large space dimension; 2) the style space is better disentangled than $\mathcal{W}^+$ space inherently; 3) Our reduced style space further improves the disentanglement capability explicitly.

Benefits of $\Delta\bm{P}$: Similarly, as shown in the last row of \Tref{tbl:ab_disentangle} (our default setting), by replacing the absolute target parameter $\bm{P}_t$ with the relative change $\Delta\bm{P}$ as the input of the latent manipulation network, the disentanglement performance significantly improves by reducing from $69.54\%$ to $59.36\%$.
Similar improvement is observed from $72.33\%$ of $\mathcal{S}$ \& $\bm{P}$ to $65.26\%$ of $\mathcal{S}$ \& $\Delta \bm{P}$.
One possible explanation is that since $\bm{P}_t$ contains other attributes' information, forcing the manipulation network to deliberately ignore such information (if given) will increase the learning difficulty of one specific attribute. In contrast, using $\Delta \bm{P}$ is equivalent to explicitly telling the network which attribute should be changed and which attribute should not be changed. Intuitively, it is also more natural as the design motive of the manipulation network is just to build the relationship between the change in the latent space and the change in the 3DMM space.

Besides the above quantitative results, we also provide some representative visual results in Fig.~\ref{fig:ab_images}. Compared to our default setting, the three remaining settings either produce serious artifacts or incur unexpected attribute change (\textit{e.g.} identity) when editing one specific attribute. Taking the ``right light'' editing case of the woman image as an example, the ``identity'' attribute seems changed in all the baseline settings. This is consistent with the conclusion drawn from the quantitative evaluations.








\subsection{Ablation Study for Latent-Consistent Finetuning}
Latent-consistent finetuning between $\text{G}$ for human faces and $\text{G}^*$ for out-of-domain faces (Fig.~\ref{fig:latent_consist}) is the key for our cross-domain editing. To evaluate its effectiveness, we conduct an ablation study. Given a human face image, we generate the corresponding out-of-domain images using the same latent code and manipulate them with the same $\Delta\bm{w}$.
With our finetuning, the out-of-domain images (Fig.~\ref{fig:latent_consist}(b)) share similar semantic attributes, such as the shape and the expression, with the human faces (Fig.~\ref{fig:latent_consist}(a)). Furthermore, when editing the attributes like the illumination, the lighting change on the human faces is preserved in the generated out-of-domain images. In contrast, without our finetuning, inconsistent semantics and editings are observed (Fig.~\ref{fig:latent_consist}(c)), even with the same manipulation offset $\Delta\bm{w}$. The quantitative evaluation on a large and uncurated set of faces can be found in the supplemental material.

\subsection{Comparisons}
To demonstrate our advantages in cross-domain face manipulation, we conduct comparisons with different types of representative baselines, including 2D warping based methods: Moving-Least-Squares (MLS) deformation~\cite{DBLP:journals/tog/SchaeferMW06} and First Order motion model~\cite{DBLP:conf/nips/SiarohinLT0S19}, latent space manipulation based methods: GANSpace~\cite{DBLP:conf/nips/HarkonenHLP20} and InterFaceGAN~\cite{DBLP:conf/cvpr/ShenGTZ20}, and 3DMM guided methods: StyleRig~\cite{DBLP:conf/cvpr/TewariEBBSPZT20} and PIE~\cite{DBLP:journals/tog/TewariE0BSPZT20}. For MLS, we guide the warping with target facial landmarks that are manually labeled for the best accuracy.
For First Order, we construct video sequences from target images to drive the out-of-domain images. For all latent manipulation methods, the pretrained StyleGAN2 model is finetuned on the same out-of-domain datasets as ours.

Comparisons with 2D Warping Based Methods: We first compare our method with 2D warping based methods. In Fig.~\ref{fig:cmp_warp}, we show the editing results of changes in terms of shape, pose, and expression, respectively. 
Since purely pixel-level warping methods like MLS cannot synthesize originally occluded regions, it fails at manipulations such as mouth opening. While the First Order motion model can handle occlusion quite well, it suffers from the same issue as MLS that 2D warping field interpolated from sparse landmarks cannot well represent 3D deformation such as head rotation. In contrast, by leveraging the generative nature of GAN and the 3D prior guidance, our method achieves higher fidelity and controllability without these artifacts.

\begin{figure*}[ht!]
      \centering
      \setlength{\tabcolsep}{0\linewidth}
      \begin{tabular}{C{0.096\linewidth}C{0.012\linewidth}C{0.096\linewidth}C{0.096\linewidth}C{0.096\linewidth}C{0.012\linewidth}C{0.096\linewidth}C{0.096\linewidth}C{0.096\linewidth}C{0.012\linewidth}C{0.096\linewidth}C{0.096\linewidth}C{0.096\linewidth}}
      &&\multicolumn{3}{c}{pose}&&\multicolumn{3}{c}{illumination}&&\multicolumn{3}{c}{expression}\\
      \multicolumn{13}{c}{\includegraphics[width=\linewidth]{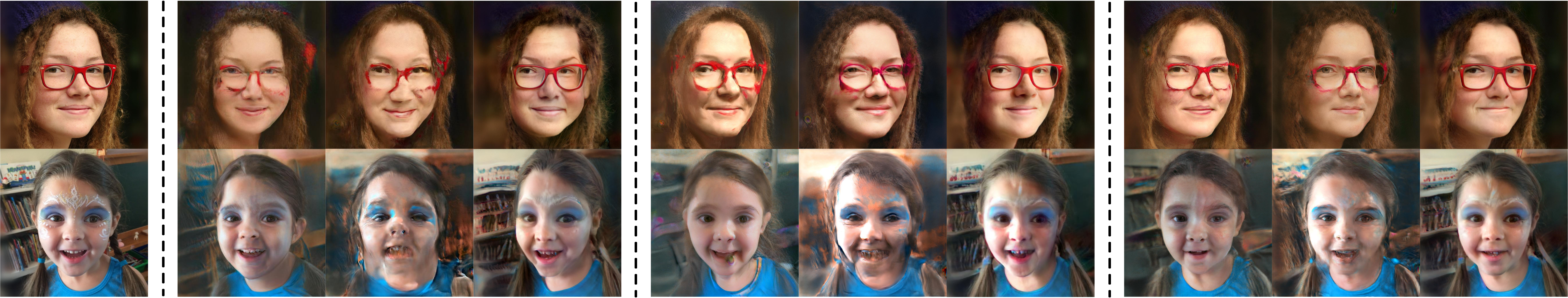}}\\
      original&&\ PIE&StyleRig&ours&&\ PIE&\ StyleRig&ours&&\ PIE&\ \ StyleRig&ours\\
      \end{tabular}
      \caption{Comparison with 3DMM guided methods StyleRig and PIE. As they do not support out-of-domain face editing, we only show the editing results for the real human faces. It shows that our method can disentangle different attributes better and produce more controllable editing results with higher quality. Results of PIE and StyleRig are directly fetched from the authors' project webpage. We show different editing directions for each attribute.}
      \label{figure:cmp3d}
\end{figure*}

Comparisons with Latent Space Manipulation Methods:
We then compare our method with GANSpace and InterFaceGAN that perform semantic image control via latent space manipulation in Fig.~\ref{figure:cmpgan}.
Since GANSpace is an unsupervised method without explicit supervision of the semantic attributes, it is difficult to locate the corresponding latent variable for a given attribute. Also, it is not guaranteed that all the factors of interest will be disentangled, \textit{e.g.}, the hair is also changed when ``opening mouth''. For InterFaceGAN, as their original version only supports human faces, we combine it with our cross-domain adaptation method when applied to out-of-domain faces. Compared to GANSpace, InterFaceGAN allows explicit controls with supervision, but the required binary labeling hardly exists for some attributes, thus not supporting the controllable editing for such attributes (\textit{e.g.} illumination, shape). In contrast, our method can disentangle different attributes much better and achieve better editing quality. 

Comparisons with 3DMM Guided Methods:
Finally, we compare our method with StyleRig and PIE, which also leverage the 3DMM guidance for face manipulation. However, due to the lack of 3DMM parametric space for out-of-domain faces, they do not support out-of-domain face editing inherently. Therefore, we only make the comparison on human face images in Fig.~\ref{figure:cmp3d}. All the results of StyleRig and PIE are directly fetched from the authors' project page.
Without our reduced latent space and disentangled attribute-latent mapping, on the one hand, they cannot manipulate identity-related attributes such as shape and albedo.
On the other hand, they often fail at editing a single attribute without affecting other content or manipulating multiple attributes simultaneously. For example, when changing the expression in the first case, the glasses are a little twisted. Similarly, when changing the expression in the second case, the albedo and background are also significantly changed with obvious artifacts. 

\subsection{User Study}

\begin{table}[t]
\caption{User study results. The values for each attribute represent the user preference rate (\%, the higher, the better) by comparing the editing results among different methods. ``-'' means that the method does not support editing this attribute or the editing direction is not available in their paper.}
\label{tbl:ab_userstudy}
\tabcolsep=5.0pt
\begin{tabular}{C{60pt}|C{30pt}C{30pt}C{30pt}C{30pt}}
\toprule
Method & shape & exp. & illum. & pose \\ 
\toprule
GANSpace & 10.57 & 0.86 & 13.71 & 0.57  \\
First Order & - & 4.29 & - & 2.86   \\
InterFaceGAN & - & 2.86 & - & 2.00    \\
MLS & 0.57 & 0.57    & - & -     \\
Ours & \textbf{88.86} & \textbf{91.42} & \textbf{86.29} & \textbf{94.57}    \\
\bottomrule
\end{tabular}
\end{table}

To quantitatively compare our method and baseline methods in terms of the out-of-domain face editing quality, we conduct a user study and let users choose their preferred results. Since StyleRig and PIE only support real human face image editing, we do not include them here. Similarly, as changing the albedo attribute is not supported in the remaining baseline methods, only the controllable editing results for shape, expression, illumination, and pose are compared. If the baseline method does not support editing one specific attribute, it will be ignored in that corresponding comparison. Specifically, we randomly choose $10$ cases for each domain, and thus obtain $50$ different cases in total for $5$ domains.
Then we use different methods to edit the target attribute while keeping other attributes unchanged. For each case, we show the edited results together with the original image to total $35$ participants and ask them the question ``Which result is the best for the ``X change'' while keeping other attributes in the original image intact, including the identity, hair style and other face details?'' Here, ``X change'' indicates different attribute editing operation, like ``being fat'' and ``turn right''. The final preference rate is defined as the average percentage of one specific method selected as the best. As shown in \Tref{tbl:ab_userstudy}, comparing to all the baseline methods, the users prefer the editing results by our method by a large margin, which is consistent with the conclusion drawn by the above visual comparisons.


\section{Results}

\begin{figure}[t]
      \begin{subfigure}{0.5\textwidth}
      \centering
      \setlength{\tabcolsep}{0\linewidth}
      \begin{tabular}{C{0.198\linewidth}C{0.01025\linewidth}C{0.198\linewidth}C{0.198\linewidth}C{0.198\linewidth}C{0.198\linewidth}}
      \multicolumn{6}{c}{\includegraphics[width=0.99825\linewidth]{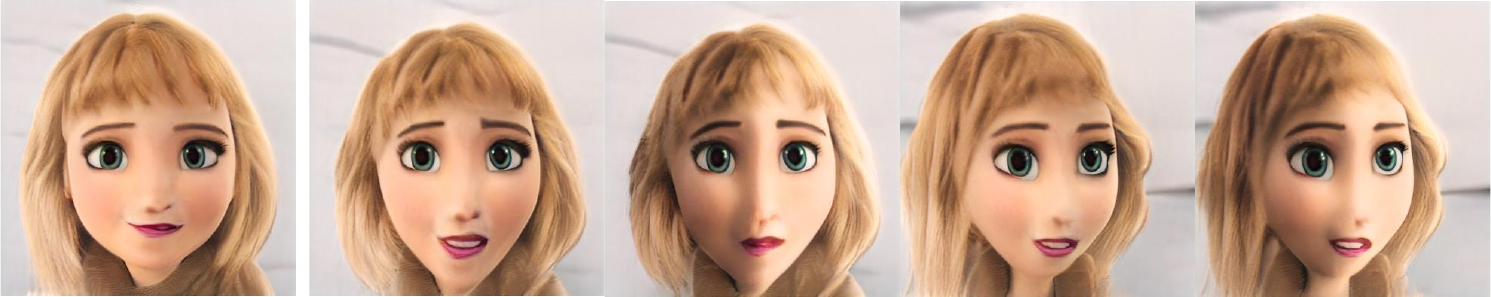}}\\
      original&&S+E&S+I&S+E+P&S+E+P+I\\
      \end{tabular}
      \caption{\textcolor{black}{Manipulations of shape (thin face), expression (open mouth), illumination (right light), and pose (turn right)}}
      \end{subfigure}
      \begin{subfigure}{0.5\textwidth}
      \centering
      \setlength{\tabcolsep}{0\linewidth}
      \begin{tabular}{C{0.198\linewidth}C{0.01025\linewidth}C{0.198\linewidth}C{0.198\linewidth}C{0.198\linewidth}C{0.198\linewidth}}
      \multicolumn{6}{c}{\includegraphics[width=0.99825\linewidth]{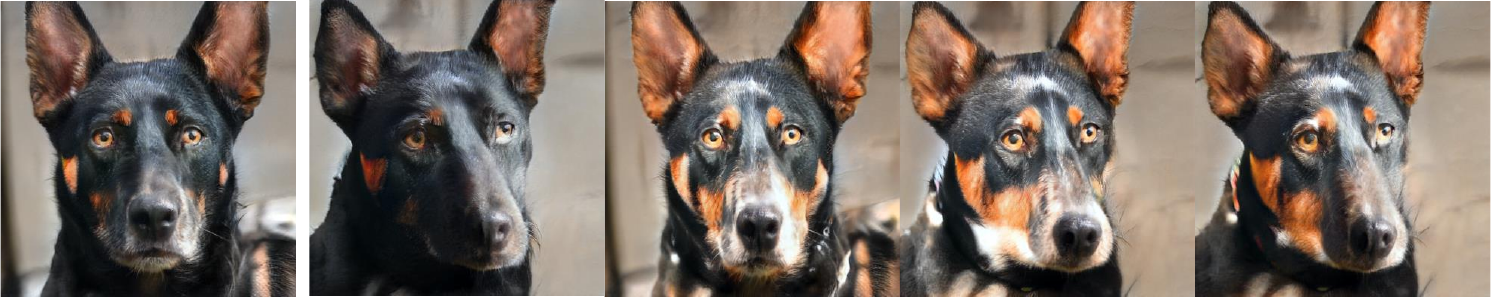}}\\
      original&&P+I&S+A&S+A+P&S+A+P+I\\
      \end{tabular}
      \caption{\textcolor{black}{Manipulations of shape (thin face), albedo (turn black), illumination (right light), and pose (turn right)}}
      \end{subfigure}
      \begin{subfigure}{0.5\textwidth}
      \centering
      \setlength{\tabcolsep}{0\linewidth}
      \begin{tabular}{C{0.198\linewidth}C{0.01025\linewidth}C{0.198\linewidth}C{0.198\linewidth}C{0.198\linewidth}C{0.198\linewidth}}
      \multicolumn{6}{c}{\includegraphics[width=0.99825\linewidth]{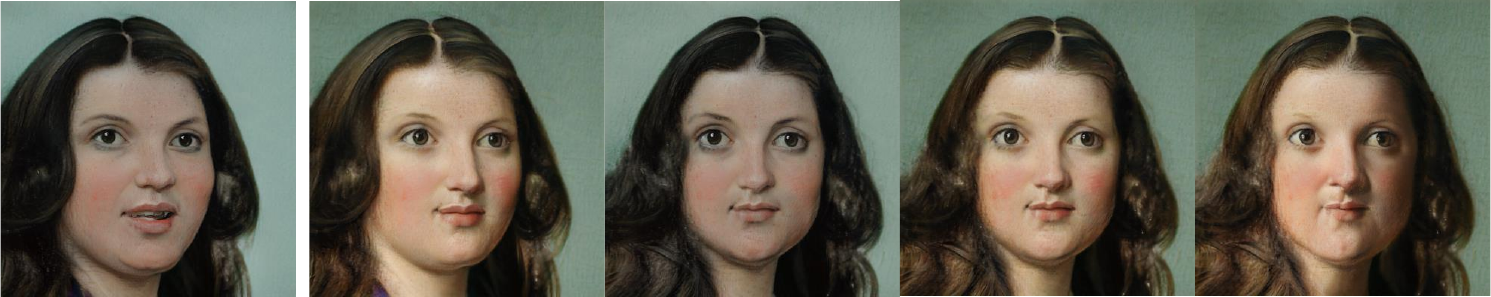}}\\
      original&&E+A&E+P&E+P+A&E+P+A+I\\
      \end{tabular}
      \caption{\textcolor{black}{Manipulations of albedo (turn white), expression (close mouth), illumination (left light), and pose (turn left)}}
      \end{subfigure}
      \caption{Multi-attribute manipulation results for shape (S), expression (E), pose (P), illumination (I), and albedo (A). Our method can manipulate multiple attributes simultaneously and generate visually plausible results. }
      \label{fig:multi_att_rst}
\end{figure}

\begin{figure}[t]
      \centering
      \setlength{\tabcolsep}{0\linewidth}
      \begin{tabular}{C{0.2\linewidth}C{0.2\linewidth}C{0.2\linewidth}C{0.2\linewidth}C{0.2\linewidth}}
      \multicolumn{5}{c}{\includegraphics[width=\linewidth]{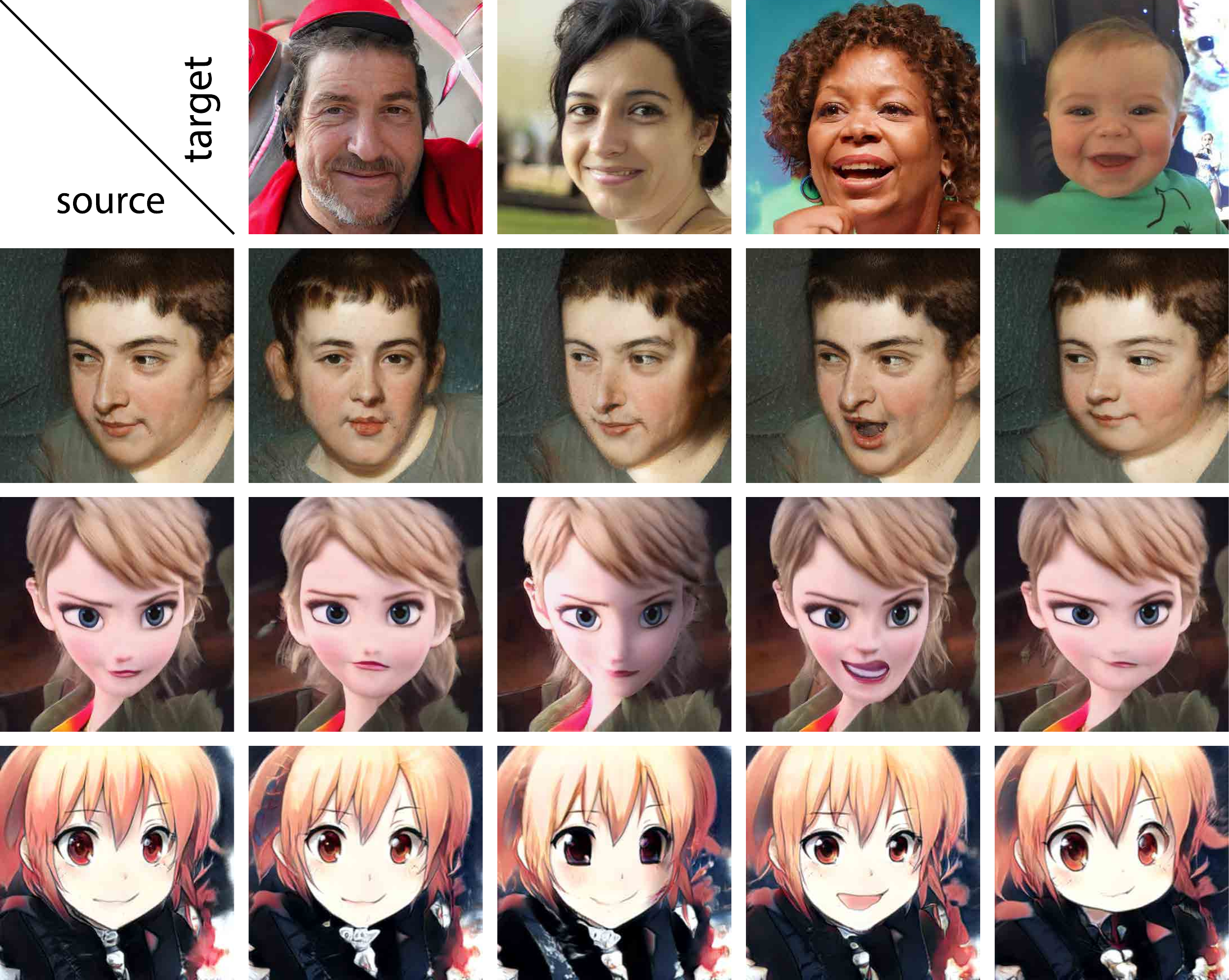}}\\
      &pose&illum.&exp.&shape\\
      \end{tabular}
      \caption{Visual results to show the cross-domain generalization ability. Given the same reference image, we can edit different out-of-domain images by changing one specific attribute while keeping others unchanged.}
      \label{fig:cross_domain}
\end{figure}

\subsection{Multi-Attribute Manipulation}
While all results in the experiments above are obtained by editing a single attribute while keeping others unchanged, our method actually supports simultaneous editing of multiple attributes. Specifically, we directly add all the latent code changes $\Delta\bm{w}^i$ onto the original code $\bm{w}_s$ for each involved attribute $i$, and then feed the edited latent code into the finetuned StyleGAN2 to generate the final result. In Fig.~\ref{fig:multi_att_rst}, we show three representative cases of editing two to four attributes. However, the baseline methods of StyleRig and PIE only support editing multiple attributes sequentially since they cannot disentangle each attribute very well and the latent code changes for different attributes interfere with each other. In contrast, our disentangled attribute-latent mapping significantly reduces the overlapping between attributes, making it possible to achieve multi-attribute manipulation without introducing significant interference.

\begin{figure*}[t]
      \centering
      \includegraphics[width=\linewidth]{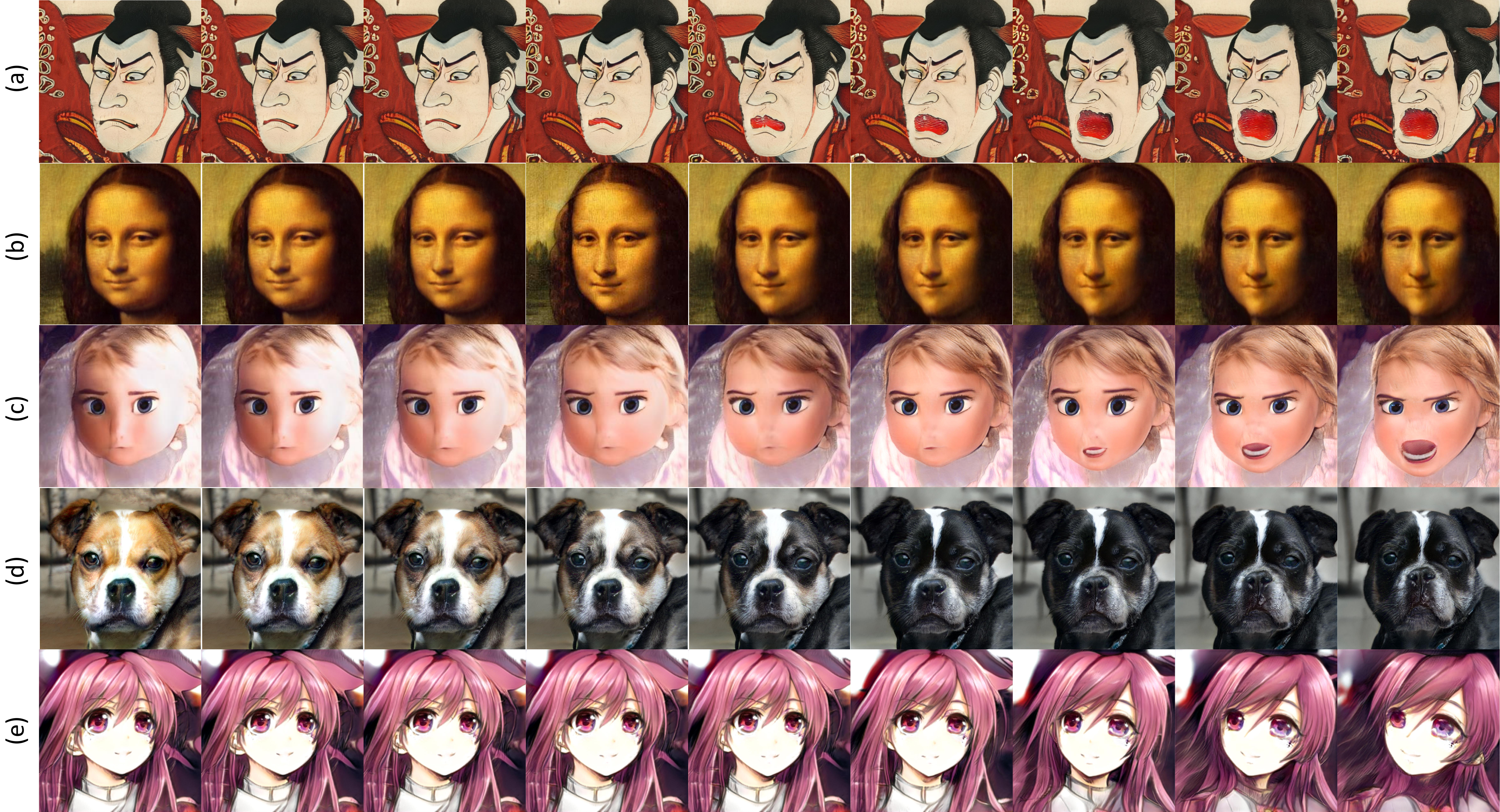}
      \caption{Examples of continuous manipulation on the attributes, including shape, expression, pose, illumination, and albedo. (a): Editing the expression. (b): Editing the expression and horizontal pose. (c): Editing the illumination and expression. (d): Editing the albedo and vertical pose. (e): Editing the pose. In each row, the leftmost image is the input, and the other images are the continuous manipulated results. }
      \label{fig:continuous_mani}
\end{figure*}

\subsection{Continuous Manipulation} Our method naturally supports continuous manipulation for changing the editing strengths of one specific attribute and changing one specific attribute to another attribute in a smooth way. Fig.~\ref{fig:continuous_mani} shows five such continuous editing cases. Taking the third baby girl as an example, we first gradually change the light from right to middle, and then change the mouth from close to open. This continuous editing function not only demonstrates the great disentanglement of our method but also potentially enables one new effect, \textit{i.e.}, bringing a still out-of-domain face image to life. Our method also supports multi-attribute continuous manipulation, which can be found in the supplementary video.

\subsection{Cross-Domain Generalization}
In Fig.~\ref{fig:cross_domain}, we demonstrate the cross-domain generalization ability of our method. For the source images from different out-of-domain face domains, we aim to change the same attribute of all the images by adopting the same reference face image. It shows that, for each attribute specified by the human face, our method can consistently follow the editing direction and achieve the corresponding editing effect. This indeed echoes our main contribution, \textit{i.e.}, disentangled out-of-domain face manipulation via bridging the discrepancy between real human face and out-of-domain face in the latent space.

\section{Conclusion}
We present the first approach for semantic attribute manipulation of out-of-domain faces by adopting 3DMM of human faces as the proxy. To this end, we devise a cross-domain adaptation method that bridges domain discrepancies and allows 3DMM parameter edits on the human face to be faithfully reflected on the out-of-domain face image. In addition, we propose a reduced latent space and disentangled attribute-latent mapping to guarantee disentangled and precise controls for each semantic attribute. With our approach, there is no need to build 3DMM for a specific out-of-domain face domain, and intuitive parameter editing, including the head pose,  facial shape, expression, albedo, and illumination, is well supported for arbitrary out-of-domain faces.  The visually pleasant quality and user-friendly control demonstrate the great potential of our method for many exciting applications in the areas of design, cartoons, animations, and games.

\begin{figure}[t]
      \centering
      \setlength{\tabcolsep}{0\linewidth}
      \begin{tabular}{C{0.247\linewidth}C{0.01025\linewidth}C{0.247\linewidth}C{0.247\linewidth}C{0.247\linewidth}}
      \multicolumn{5}{c}{\includegraphics[width=\linewidth]{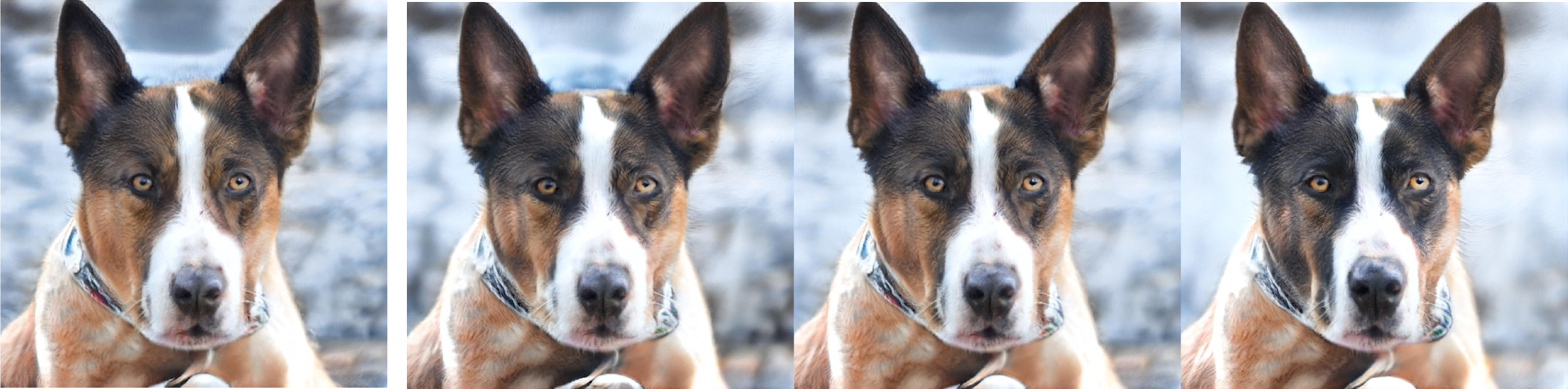}}\\
      original&&smile&open mouth&close mouth\\
      \multicolumn{5}{c}{\includegraphics[width=\linewidth]{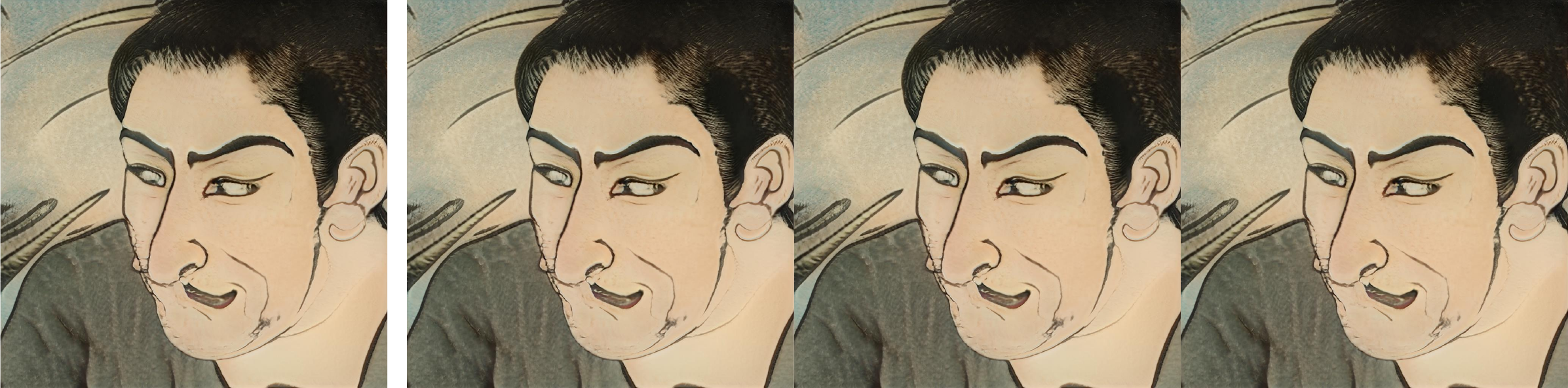}}\\
      original&&left light&right light&top light\\
      \end{tabular}
      \caption{Failure cases of expression and illumination editing on the AFHQ Dog and Ukiyo-e Face domains, respectively. The reason for the failure is that the datasets used for finetuning the StyleGAN2 do not contain enough variations for these attributes.}
      \label{fig:failure_case}
\end{figure}

Since our approach is one of the first steps towards 3DMM-based manipulation of out-of-domain faces, there is still room for further improvement. First, some fine-grained edits beyond the expressivity of 3DMM, like wrinkling nose or wearing glasses, are not supported by our method. This might be improved by adopting more expressive parametric face models. Second, some 3DMM attribute edits may not be effective in some specific domains. As shown in \Fref{fig:failure_case}, it is hard to open the mouth of a dog or to change the illumination of a Ukiyo-e face since their datasets do not contain such variations. Third, if our method is directly applied to video frames, some popping artifacts will be noticeable. This can be alleviated by adding temporal coherency loss~\cite{DBLP:conf/iccv/ChenLYYH17} when training our disentangled attribute-latent mapping network. Moreover, our method needs to finetune the StyleGAN2 in a specific domain before manipulating images of that domain. To increase the scalability of our method, how to fast update a StyleGAN2 model with small-scale data or even a single shot would be a worth-exploring direction in the future.
\textcolor{black}{Compared to StyleRig \cite{DBLP:conf/cvpr/TewariEBBSPZT20} that is unable to manipulate albedo, our method enables albedo manipulation by the disentangled attribute latent mapping. However, we admit that some results of albedo and illumination manipulations are not perfect due to the limited ability in modeling albedo and illumination using the chosen 3DMM bases and the limited diversity in lighting conditions and skin colors of the out-of-domain face datasets. For example, in WikiArt 
dataset, most portraits are drawn under mild or bright lighting conditions, while in Danbooru2018 and Disney Face datasets, most characters have flat skin tones. We believe, as better 3DMM bases and more various data become available, our model can easily take advantage of them to further improve the performance in albedo and illumination manipulation.}

\bibliographystyle{ACM-Reference-Format}
\bibliography{ref}

\section{Supplemental Material}

\subsection{Ablation Study on Loss Functions}

\begin{figure}[t]

      \centering
      \setlength{\tabcolsep}{0\linewidth}
      \begin{tabular}{C{0.242\linewidth}C{0.03025\linewidth}C{0.242\linewidth}C{0.242\linewidth}C{0.242\linewidth}}
      \multicolumn{5}{c}{\includegraphics[width=0.96\linewidth]{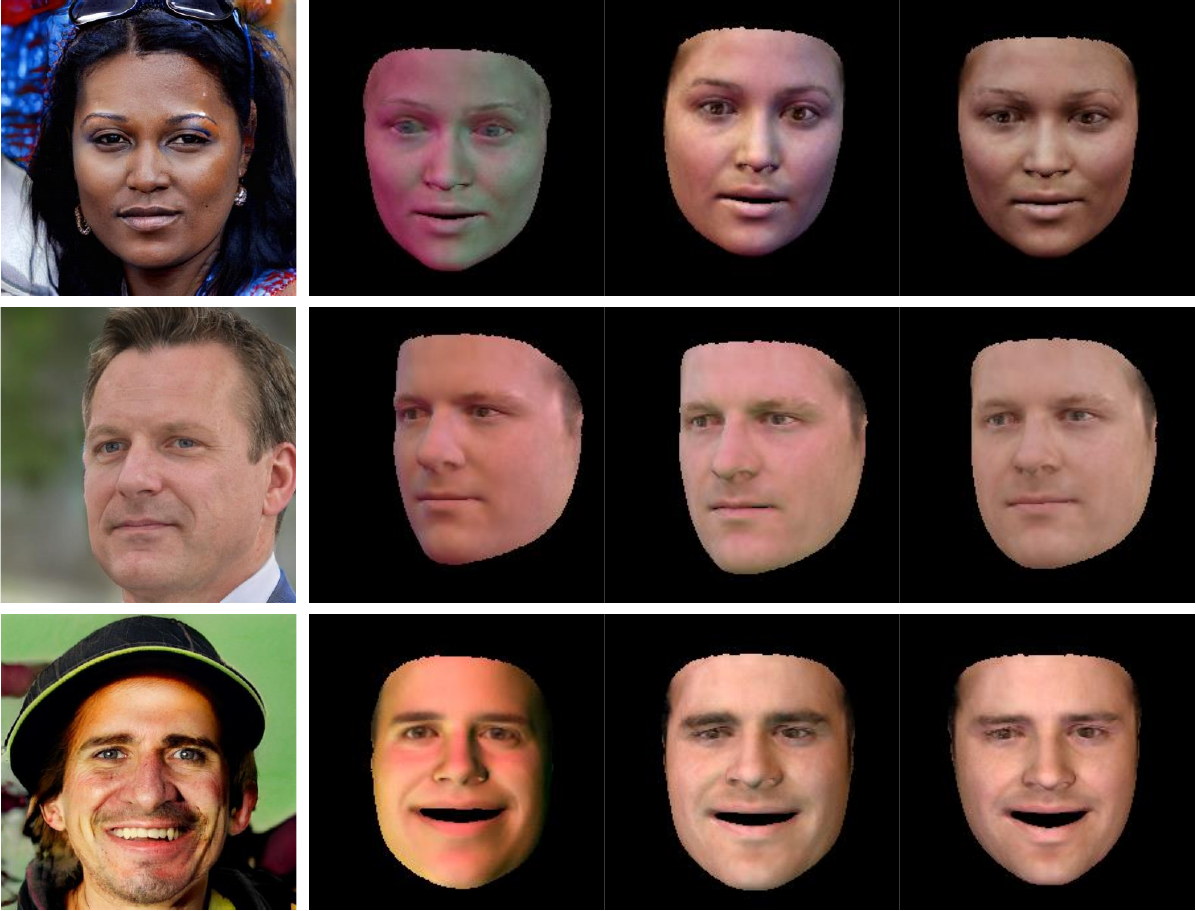}}\\
      input&&self-supervised&PDC&WPDC\\
      \end{tabular}
      
      \caption{3D face reconstruction results with different losses used in training the attribute prediction network. }
      \label{fig:ab_loss}
\end{figure}

\Fref{fig:ab_loss} shows an ablation study for training the attribute prediction network using various loss functions.
Self-supervised adopts the render loss and landmark loss, while \textcolor{black}{parameter distance cost (PDC)} does not use the weight stated in Eq.5. The self-supervised loss uses no 3D ground truth and may generate 3D meshes with inaccurate pose and albedo texture. PDC loss performs better by using ground truth pairs, but its pose and texture are still inaccurate. In contrast, \textcolor{black}{weighted parameter distance cost (WPDC)} can successfully reconstruct a visual-pleasing result by assigning a weight for each parameter dynamically adjusted during the training. Therefore, we adopt WPDC loss to train the attribute prediction network.

\subsection{User Interface}

\begin{figure}[t]
      \centering
      \includegraphics[width=\linewidth]{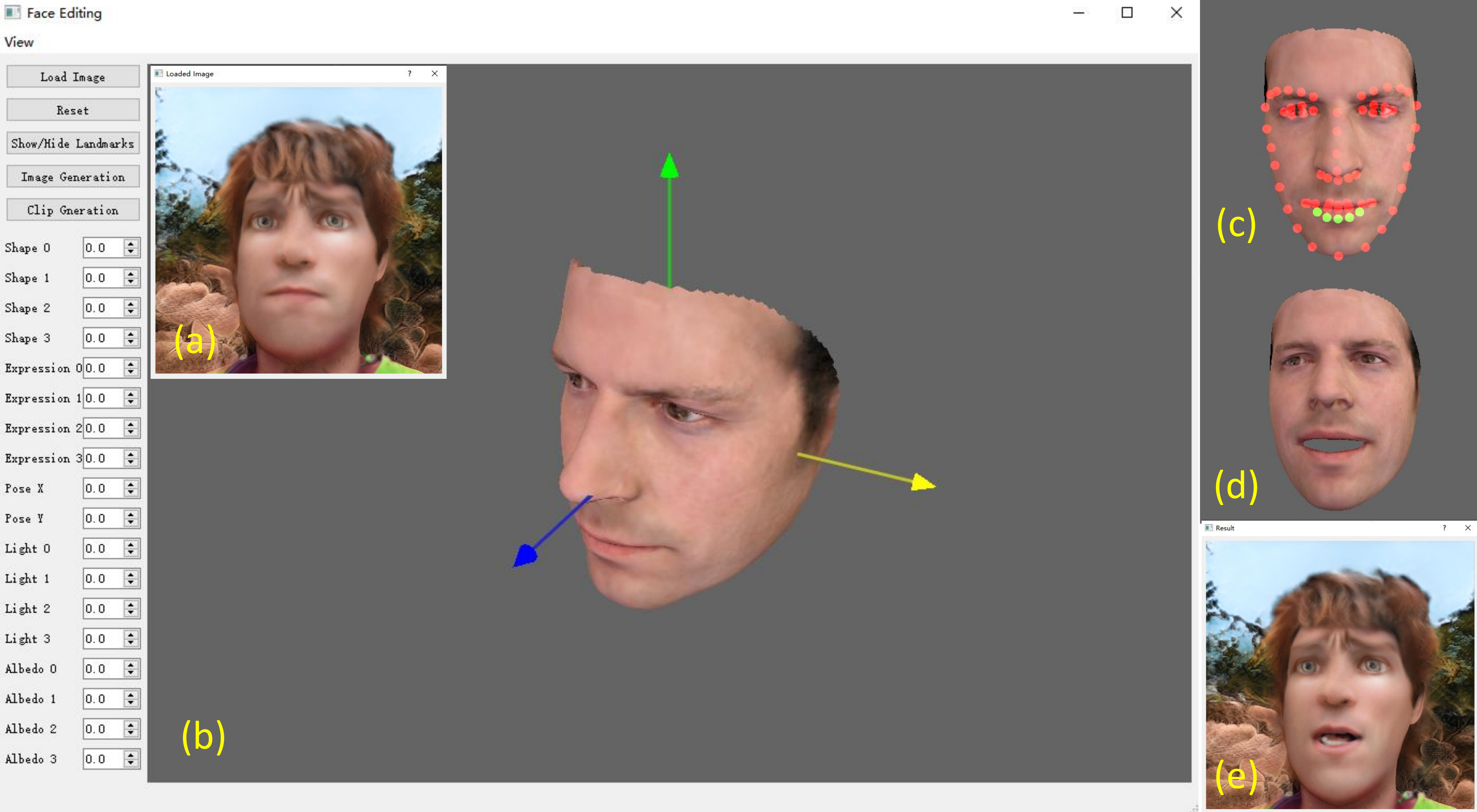}
      \caption{Our interactive cross-domain face editing system. (a) shows the input out-of-domain face image. (b) shows the reconstructed 3D mesh corresponding to the input.
      (c) shows the windows that users can edit the 3D mesh of the corresponding real face in a fine-grained way by controlling the facial landmarks. (d) shows the 3D mesh editing result. (e) is the final generated result. The left panel shows the editable parameters for each facial attribute, including shape, expression, pose, illumination and albedo.}
      \label{fig:interactive_interface}
\end{figure}
To facilitate the editing process for users, we develop an interactive editing system shown in \Fref{fig:interactive_interface}. To edit one out-of-domain face, 1) the system first inverts this image into a latent code by using the finetuned StyleGAN2; 2) then map this latent code into the 3D parameters by using the attribute prediction network and show the 3D mesh in the main window; 3) the users can edit images by adjusting values for different 3DMM bases of each attribute in the left panel or directly edit the 3D geometry by dragging the 3D facial landmarks, and the backend will use the Laplacian mesh deformation algorithm to get an edited 3D mesh; 4) the edited 3D parameters will be mapped back to the latent space by using the latent manipulation network; 5) finally the system feeds the edited latent code into the finetuned StyleGAN2 to get and show the final editing result. We show some demonstrations of the interactive system in the supplementary video.

\begin{figure*}[t]
      \centering
      \setlength{\tabcolsep}{0\linewidth}
      \includegraphics[width=0.99375\linewidth]{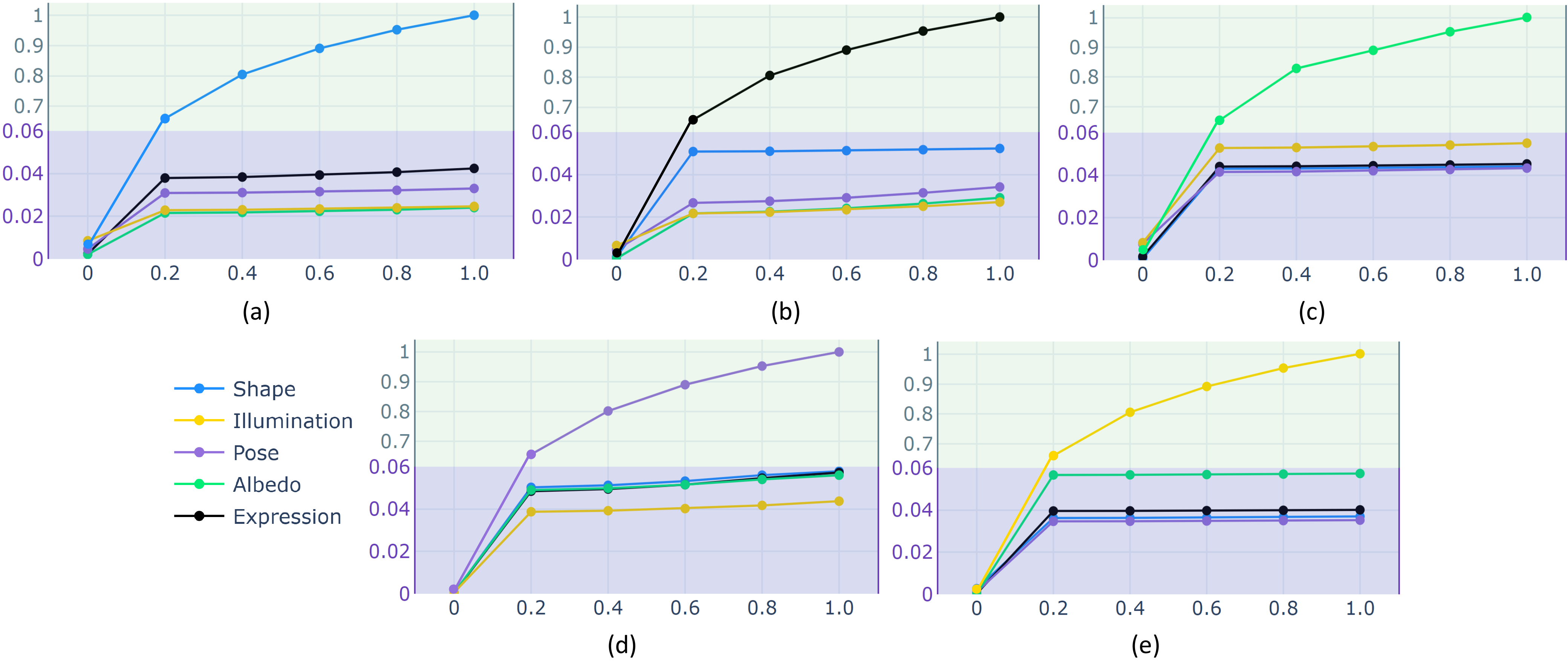}
      \caption{Attribute disentanglement evaluation. When editing one attribute, we present the average variations of each attribute in the generated results. (a): Editing shape. (b): Editing expression. (c): Editing albedo. (d): Editing pose. (e): Editing illumination. }
      \label{fig:cmp_warp}
\end{figure*}

\subsection{Attribute Disentanglement Evaluation}

In Fig.~3, we evaluate the disentanglement by editing one attribute in a range from 0 to 1.0 and then checking the parameter variations of all attributes in the generated results. Specifically, we calculate the difference between the input 3DMM parameters and these of the generated image. For better visualization, the differences are modulated with a log function, defined as $y=\log_b(1+s\times x)$. Here, the base $b$ and the scale $s$ are set as 100. 

The curves clearly show the attribute variations in the result are positively correlated with the input edit. Moreover, when we edit one attribute, \textit{e.g.}, pose, the most prominent variations in the generated results happen in the same attribute. In contrast, variations of all other attributes are close to zero (less than 0.06), which has negligible influence on the final result. This demonstrates the good ability of our method in terms of attribute disentanglement.

\subsection{Latent Space Consistency Evaluation}

\begin{figure}[t]
      \centering
      \setlength{\tabcolsep}{0\linewidth}
      \begin{tabular}{C{0.2\linewidth}C{0.2\linewidth}C{0.2\linewidth}C{0.2\linewidth}C{0.2\linewidth}}
       Prefer. rate: &\quad 87.88\%& \ 3.98\%&5.68\%&2.46\%\\
      \multicolumn{5}{c}{\includegraphics[width=\linewidth]{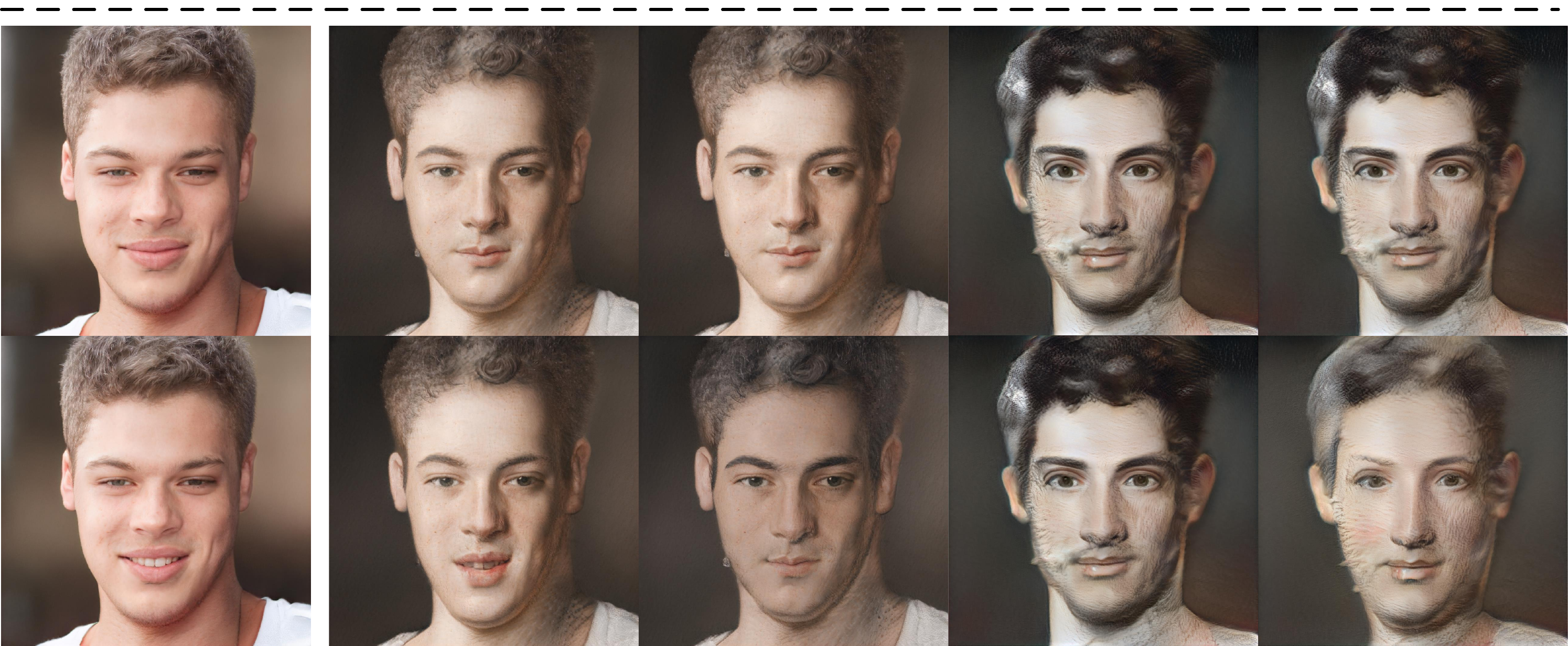}}\\
      (a)&(b)&(c)&(d)&(e)\\
      \end{tabular}
      \caption{One sample in our user study for latent space consistency. 
      (a): the image pair generated using the latent code $\bm{w}$ and $\bm{w}+\bm{\Delta w}$ via the real face generator $\text{G}$. 
      (b): the same as (a) but uses $\text{G}^*$ as the generator.  (c): replacing $\bm{\Delta w}$ in (b) with a randomly sampled ${\bm{\Delta w}}'$. (d) and (e): the same as (b) and (c) but avoid using our latent-consistent finetuning. We also report the user preference rate (\%, the higher, the better) of the user study for latent space consistency in the first row.  }
      \label{fig:ablation_consist}
\end{figure}

To quantitatively evaluate the latent space alignment between two styleGAN models: $\text{G}$ for human faces and $\text{G}^*$ for out-domain faces, we conduct a user study. Given an input real human face image with its latent code $\bm{w}$, we randomly sample a meaningful $\bm{\Delta w}$ in our reduced style space and edit the input image using this $\bm{\Delta w}$ to obtain the edited image, as shown in \Fref{fig:ablation_consist}(a). Here, $\text{G}$ is adopted as the generator.
Then $\bm{w}$ and $\bm{\Delta w}$ are used to produce an image pair in out-of-face domain, including an image generated via $\bm{w}$ and an edited image using the $\bm{\Delta w}$, as shown in \Fref{fig:ablation_consist}(b). We then generate another image pair in \Fref{fig:ablation_consist}(c) using the same latent code $\bm{w}$ but a different randomly sampled ${\bm{\Delta w}}'$. Images from \Fref{fig:ablation_consist}(b) and \Fref{fig:ablation_consist}(c) are generated using $\text{G}^*$ with our latent-consistent finetuning. Next, we present another two image pairs in \Fref{fig:ablation_consist}(d) and \Fref{fig:ablation_consist}(e) produced by using the same generation strategy as \Fref{fig:ablation_consist}(b) and \Fref{fig:ablation_consist}(c), but images from \Fref{fig:ablation_consist} (d) and \Fref{fig:ablation_consist}(e) are generated without our latent-consistent finetuning. 

Finally, 23 such cases are tested in our user study (which can be found in the supplemental web page).
For each case, one real human face pair and four out-domain pairs are presented to the user. Unlimited time is given for the user to choose one image pair from four that 1) is most semantically similar to the real human face pair and 2) shares the same editing direction with the real human face pair. We collect valid responses from 24 users. The final preference rate is defined as the percentage of a specific method to be selected as the best. As shown in \Fref{fig:ablation_consist}, our (b) outperforms other settings by a large margin, which demonstrates that our method can preserve semantics and enforce the same editing direction between the latent space of $\text{G}$ and $\text{G}^*$. These are achieved by the successful latent space alignment using our latent-consistent finetuning.

\begin{figure*}[t]
      \centering
      \setlength{\tabcolsep}{0\linewidth}
      \begin{tabular}{C{0.0909\linewidth}C{0.0909\linewidth}C{0.0909\linewidth}C{0.0909\linewidth}C{0.0909\linewidth}C{0.0909\linewidth}C{0.0909\linewidth}C{0.0909\linewidth}C{0.0909\linewidth}C{0.0909\linewidth}C{0.0909\linewidth}}
      \multicolumn{11}{c}{\includegraphics[width=0.99375\linewidth]{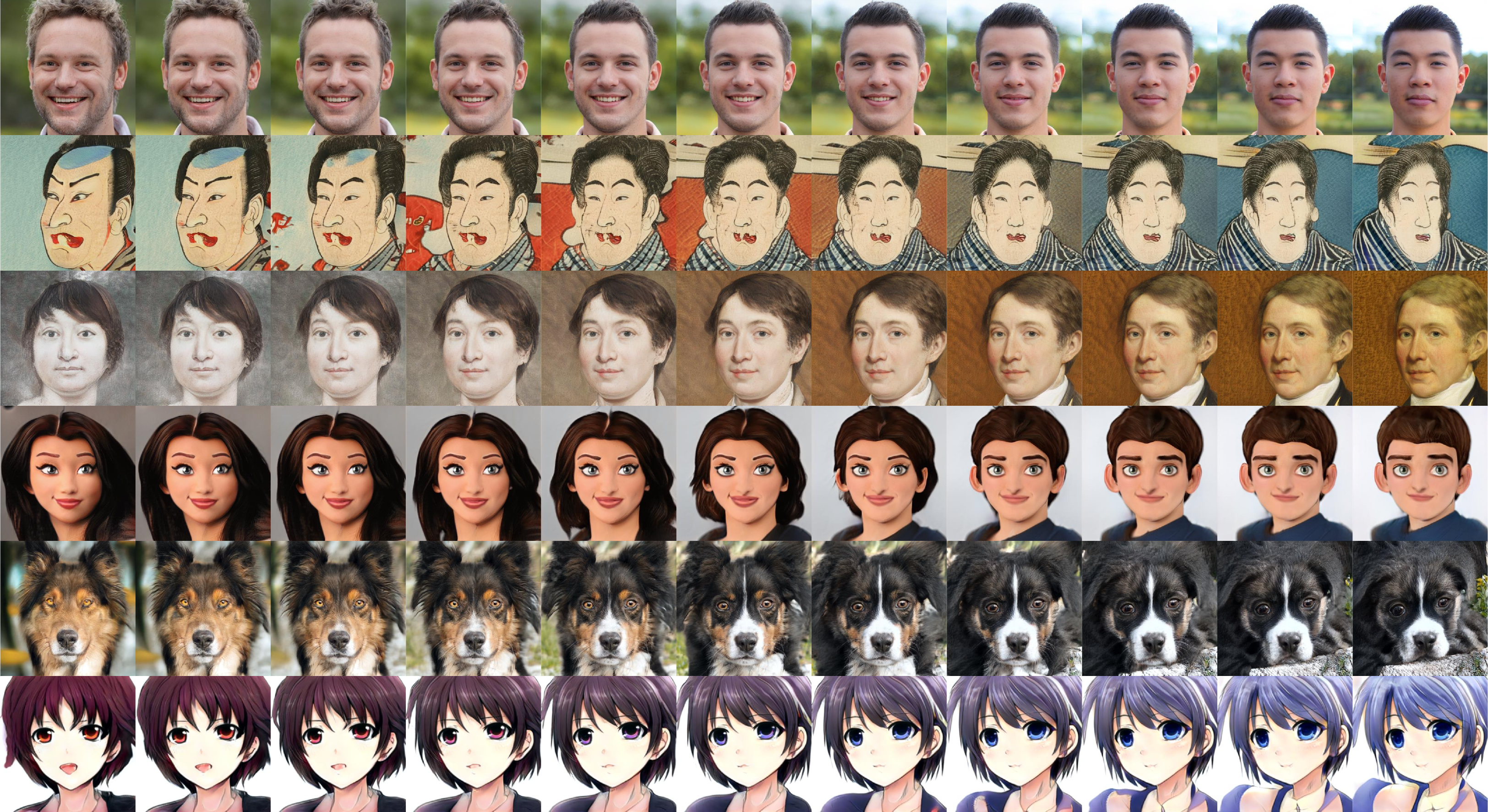}}\\
      0&0.1&0.2&0.3&0.4&0.5&0.6&0.7&0.8&0.9&1.0\\
      \end{tabular}
      \caption{Interpolations in the reduced style space of the finetuned model $\text{G}^*$. The leftmost and rightmost ones are the input images. And interpolation ratios are shown at the bottom.}
      \label{fig:interpolation}
\end{figure*}

\subsection{Latent Interpolation of Finetuned Models}

As shown in \Fref{fig:interpolation}, after our latent-consistent finetuning, the finetuned model $\text{G}^*$ still supports interpolation between two latent codes $w_s$ and $w_t$ randomly sampled from the reduced style space. Given an interpolation ratio $r$, the interpolated latent code $w_{interp}$ is defined as $w_{interp}=w_t\times r + w_s\times (1-r)$. Here $r$ ranges from 0 to 1.0 with a step 0.1. With this $w_{interp}$, we can generate the interpolated image.

\end{document}